\documentclass[a4paper,11pt,twoside,openright,final]{memoir} 
\setlrmarginsandblock{3.5cm}{2.5cm}{*}%
\setulmarginsandblock{2.5cm}{2.5cm}{*}%
\checkandfixthelayout%

\aliaspagestyle{chapter}{empty}
\aliaspagestyle{part}{empty}
\setpnumwidth{2.5em}
\setrmarg{3.5em}

\usepackage[
  plainpages=false,
  pdfpagelabels,
  pdftex,
  unicode,
  colorlinks,
  breaklinks=true,
  linkcolor=black,
  citecolor=black,
  urlcolor=black,
  pdfauthor={Fabio Ferreira},
  pdftitle={Leveraging Meta-Learning and Synthetic Data for Automated Pretraining and Finetuning in Machine Learning},
  pdfsubject={Dissertation},
  pdfkeywords={thesis, dynamic configuration, reinforcement learning, contextual reinforcement learning, algorithm configuration, dynamic algorithm configuration}
]{hyperref}

\usepackage[final]{microtype}
\usepackage[utf8]{inputenc}
\usepackage[T1]{fontenc}
\usepackage{lmodern}
\usepackage{csquotes}
\usepackage[ngerman,english]{babel}
\usepackage[
backend=biber,
natbib=true,
style=authoryear-comp,
maxbibnames=99,
maxcitenames=2,
uniquelist=false,
sorting=nyt,
mincrossrefs=99,
dashed=false,
backref=true,
]{biblatex}
\BiblatexSplitbibDefernumbersWarningOff
\usepackage{amsmath,amssymb,amsthm,amsfonts}
\usepackage{mathabx} 
\usepackage[mathscr]{eucal}
\usepackage{bm}
\usepackage{xurl}
\usepackage{mdframed}

\newcommand{\citeaddendum}{}

\makeatletter
\DeclareCiteCommand{\fullcite}
{\defcounter{maxnames}{\blx@maxbibnames}%
  \usebibmacro{prenote}}
{\usedriver
  {\DeclareNameAlias{sortname}{default}}
  {\thefield{entrytype}}.\citeaddendum{}}
{\multicitedelim}
{\usebibmacro{postnote}}
\makeatother

\makeatletter
\DeclareMathSizes{\@xipt}{11}{9}{8}
\makeatother

\usepackage{etoolbox}
\usepackage{breqn} 
\usepackage{hyphenat} 
\usepackage{graphicx}
\usepackage{charter} 
\usepackage{multirow}
\usepackage{dirtytalk} 
\usepackage{paralist} 
\usepackage{nicefrac}
\usepackage{subfig}
\usepackage{array}
\usepackage[noabbrev]{cleveref}
\usepackage{pstricks}

\usepackage{tcolorbox} 

\usepackage[noend]{algpseudocode}
\usepackage{algorithm,algorithmicx}
\usepackage{pdfpages} 

\epigraphfontsize{\normalsize\itshape}
\usepackage{tikz}
\usetikzlibrary{arrows}
\usetikzlibrary{arrows.meta}
\usetikzlibrary{graphs,graphs.standard,calc}
\usetikzlibrary{shapes,decorations}
\tikzstyle{cross}=[cross out, draw=black, minimum size=2*(#1-\pgflinewidth), inner sep=0pt, outer sep=0pt]
\usepackage{varwidth}
\usepackage{parskip}
\usepackage{fontawesome}
\usepackage{verbatim}
\usepackage{pgfplots}
\pgfplotsset{compat=1.18}

\usepackage{lipsum}
\usepackage{tipa}

\usepackage[colorinlistoftodos,color=yellow]{todonotes}

\usepackage{nameref}
\usepackage{mwe}
\usepackage{wrapfig}
\usepackage{tikzducks}
\chapterstyle{madsen}
\setsecnumdepth{subsection} 
\AtEveryBibitem{%
  \ifcategory{exclude}
    {\csappto{blx@bibnote}{\DeclareFieldFormat{extraalpha}{}{}}%
     \csappto{blx@bibnote}{\DeclareFieldFormat{extrayear}{}{}}}
    {}%
}

\newcommand*\cleartoleftpage{%
  \clearpage
  \ifodd\value{page}\hbox{}\newpage\fi
}
\counterwithout{footnote}{chapter}

\theoremstyle{plain}

\theoremstyle{definition}

\theoremstyle{remark}

\newcommand{\AB}[0]{André Biedenkapp}

\newcommand{\highlightfullcite}[1]{%
\begin{mdframed}[backgroundcolor=gray!20]%
\fullcite{#1}%
\end{mdframed}%
}
\DeclareMathOperator*{\argmin}{arg\,min}

\newcommand*{\fullref}[1]{\hyperref[{#1}]{\cref*{#1}: \nameref*{#1}}} 
\newcommand*{\Fullref}[1]{\hyperref[{#1}]{\Cref*{#1}: \nameref*{#1}}} 
\newcommand*{\bfullref}[1]{\hyperref[{#1}]{\cref*{#1} (\nameref*{#1})}} 
\newcommand*{\BFullref}[1]{\hyperref[{#1}]{\Cref*{#1} (\nameref*{#1})}} 

\let\cite\citep

\definecolor{darkred}{rgb}{0.75, 0.0, 0.0}

\DeclareBibliographyCategory{exclude}

\DeclareNameFormat{highlightname}{%
  \namepartgiveni
  \ifstrequal{\namepartfamily}{Ferreira}
    {\textcolor{darkred}{\namepartfamily}}
    {\namepartfamily}
  \ifthenelse{\value{listcount}<\value{liststop}}
    {\addcomma\space}
    {}%
}

\makeatletter
\DeclareCiteCommand{\fullciteexclude}
  {%
    \defcounter{maxnames}{\blx@maxbibnames}%
    \usebibmacro{prenote}%
  }
  {%
    \usedriver
      {\DeclareNameAlias{sortname}{highlightname}}%
      {\thefield{entrytype}}%
    \citeaddendum{}%
    \addtocategory{exclude}{\thefield{entrykey}}%
  }
  {%
    \multicitedelim%
  }
  {%
    \usebibmacro{postnote}%
  }
\makeatother

\begin{document}
\begin{titlingpage}


\huge \centering{\textbf{Meta-Learning and Synthetic Data for Automated Pretraining and Finetuning
}}\\

  \vspace{2cm}
  
  \centering{\resizebox*{0.7\textwidth}{!}{
      \includegraphics{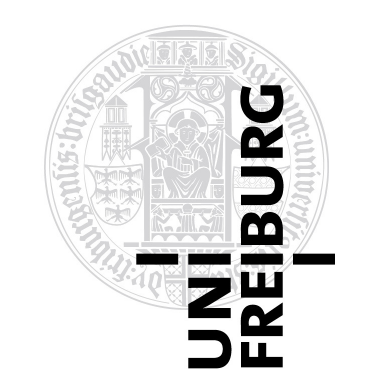}}}

  \vspace{2cm}


    \LARGE \centering{\textsf{Dissertation zur Erlangung des Doktorgrades der \\Technischen Fakult{\"a}t der Albert-Ludwigs-Universit{\"a}t Freiburg}}

  \vspace{1.5cm}

  \centering{vorgelegt von\\} 

  \vspace{0.5cm} 
  
  \centering{\textbf{Fabio Ferreira}}
  
  \vspace{0.5cm} 

  \centering{2025}

  \normalsize

  \clearpage

  \vspace*{\fill}
  \begin{flushleft}
    \noindent
    \textbf{Dean:}\\
    Prof. Dr.-Ing. Frank Balle, \emph{University of Freiburg, Germany}\\

    \bigskip

    \noindent
    \textbf{PhD advisor and first examiner:}\\
    Prof. Dr. Frank Hutter, \emph{University of Freiburg, Germany}\\

    \bigskip

    \noindent
    \textbf{Second examiner:}\\
    Prof. Dr. Josif Grabocka, \emph{University of Technology Nuremberg, Germany}\\

    \bigskip


    \bigskip

    \noindent
    \textbf{Date of defense: June 2, 2025}\\

  \end{flushleft}
\end{titlingpage}

\frontmatter
\thispagestyle{empty}
\chapter*{Abstract}
\vspace{-1cm}
\begin{flushright}
Machine Learning $\cdot$ Automated Machine Learning $\cdot$ Meta-Learning $\cdot$ Deep Learning
\end{flushright}
The growing number of pretrained models in Machine Learning (ML) presents significant challenges for practitioners. Given a new dataset, they need to determine the most suitable deep learning (DL) pipeline, consisting of the pretrained model and the hyperparameters for finetuning to it. 
Moreover, as models grow in scale, the increasing reliance on real-world data poses a bottleneck for training and requires leveraging data more effectively.

Addressing the first challenge often involves manual model selection and hyperparameter tuning, demanding expert knowledge or costly trial and error. At the same time, as models grow larger and more and more of the available human-generated data is being used for training, data augmentation and synthetic data become critical elements in today's pipelines to improve effective data use and model performance.
Automated machine learning offers a path to address these challenges but is traditionally designed for tabular data and classical ML methods. Extending these techniques to Deep Learning presents a key challenge due to high computational and data-related demands.
This dissertation adopts meta-learning to extend automated machine learning to the deep learning domain. Its contributions are structured into two clusters: 


In the first cluster, we propose empirical approaches to automate DL pipeline selection for Computer Vision tasks using prior task knowledge to learn zero-shot and few-shot surrogate models. Extending these methods to the language domain, we demonstrate their efficacy in meta-learning for finetuning large language models (LLMs) and their cross-domain applicability. Not only removing the reliance on manual selection, we empirically show that our approach generalizes to unseen datasets to suggest task-specific pipelines that can outperform finetuning large, jack-of-all-trades backbone models.

The second cluster focuses on meta-learning for data augmentation and synthetic data to enhance performance and effectively use data in pretraining and finetuning regimes. We empirically show the underestimated importance of data augmentation when using Self-Supervised Learning in the computer vision domain and, based on that, meta-learn advanced data augmentation strategies that enhance pretraining and finetuning performance. Leveraging synthetic data, we propose a framework to meta-learn neural synthetic data generators as proxies for Reinforcement Learning (RL) environments, enabling efficient and hyperparameter-agnostic agent training. Additionally, we explore using synthetic, randomly sampled data to pretrain a general simulator in an in-context learning fashion that covers multiple RL environments, allowing hyperparameter-agnostic training without real-world interactions.

Overall, this dissertation takes a step toward automated pretraining and customizable finetuning through meta-learning and empirically demonstrates how meta-learning can leverage data augmentation and synthetic data for scalable, domain-adaptable machine learning.
\thispagestyle{empty}
\selectlanguage{ngerman}

\chapter*{Zusammenfassung}
\vspace{-1cm}
\begin{flushright}
Maschinelles Lernen $\cdot$ Automatisiertes Maschinelles Lernen $\cdot$ Meta-Lernen $\cdot$  Tiefes Lernen $\cdot$ 
\end{flushright}

Die wachsende Zahl von vortrainierten Modellen im maschinellen Lernen (ML) stellt Praktiker vor große Herausforderungen. Beim Einsatz eines neuen Datensatzes müssen sie die geeigneten Deep-Learning-Pipeline bestimmen, bestehend aus dem vortrainierten Modell und den Hyperparametern für das Finetuning. Stetig wachsende Modelle erhöhen die Abhängigkeit von realen Daten, was einen Engpass für das Training darstellt und eine effektivere Datennutzung erfordert.

Die Modellauswahl und Hyperparameterabstimmung erfolgen oft manuell, was Expertenwissen oder kostenintensive Experimente verlangt. Mit stetig größer werdenden MOdellen und da immer mehr der verfügbaren, vom Menschen generierten Daten für das Training verwendet werden, werden Datenaugmentierung und synthetische Daten zu entscheidenden Elementen in den heutigen Pipelines, um die effektive Datennutzung und die Modellleistung zu verbessern. Automatisiertes maschinelles Lernen bietet einen Weg zur Bewältigung dieser Herausforderungen, ist jedoch traditionell für tabellarische Daten und klassische ML-Methoden konzipiert. Die Erweiterung dieser Techniken auf Deep Learning stellt aufgrund der hohen rechnerischen und datenbezogenen Anforderungen eine wesentliche Herausforderung dar. In dieser Dissertation wird Meta-Learning eingesetzt, um automatisiertes maschinelles Lernen auf den Bereich des Deep Learning zu erweitern. Die Beiträge dieser Arbeit sind in zwei Cluster gegliedert: 

Im ersten Cluster schlagen wir empirische Ansätze zur Automatisierung der DL-Pipeline-Auswahl für Computer-Vision-Aufgaben vor, indem wir vorheriges Aufgabenwissen zum Erlernen von Zero-Shot- und Few-Shot-Surrogatmodellen nutzen. Außerdem zeigen wie diese Methoden auf den Sprachbereich für die Feinabstimmung großer Sprachmodelle (Large Language Models; LLMs) erweitert werden können. Dabei sind wir nicht mehr auf die manuelle Auswahl angewiesen und zeigen auch empirisch, dass unser Ansatz sich auf ungesehene Datensätze generalisieren lässt, um aufgabenspezifische Pipelines vorzuschlagen, die die Feinabstimmung großer Backbone-Modelle übertreffen können.

Das zweite Cluster konzentriert sich auf Meta-Lernen für Datenaugmentierung und synthetische Daten, um die Leistung zu verbessern und Daten im Vortraining und Feinabstimmung effektiv zu nutzen. Wir zeigen empirisch die unterschätzte Bedeutung der Datenaugmentierung bei der Verwendung von selbstüberwachtem Lernen im Bereich der Computer Vision und erlernen darauf aufbauend fortgeschrittene Strategien zur Datenaugmentierung, die die Vortraining- und Feinanpassungsleistung verbessern. Außerdem demonstrieren wir, wie wir Meta-Lernen dafür nutzen können, neuronale synthetische Datengeneratoren als Stellvertreter für Reinforcement Learning (RL)-Umgebungen zu erzeugen, welche ein effizientes und hyperparameter-agnostisches Agententraining ermöglichen. Darüber hinaus untersuchen wir die Verwendung synthetischer, zufällig generierter Daten zum Vortraining eines allgemeinen Simulators, die mehrere RL-Umgebungen abdeckt und ein hyperparameter-agnostisches Training ohne Interaktionen mit der realen Welt ermöglicht.

Zusammenfassend ist diese Dissertation ein Schritt in Richtung automatisiertes Vortraining und anpassbare Feinabstimmung durch Meta-Lernen und zeigt empirisch, wie Meta-Learning dazu genutzt werden kann, effektive Datenaugmentierung und synthetische Datengenerierung für skalierbares, domänenangepasstes maschinelles Lernen zu erlangen.
\selectlanguage{english}
\thispagestyle{empty}
\chapter*{Acknowledgments}
A PhD journey stands and falls with the people who cultivate supportive and inspiring environments where one can feel comfortable and grow. Many of those who supported me in my career I was fortunate to meet early on.\\

My path as a researcher began during my fifth semester as a Computer Science undergraduate in 2015, when \emph{Prof. Astrid Laubenheimer's} Computer Vision course sparked my interest in Machine Learning (ML). The class focused on manually engineering feature extraction kernels, but it was also a time when Deep Learning (DL) was on the rise. Fascinated by the idea that DL could automate kernel learning, I decided to focus my studies entirely on ML and DL. I went on to work as a student researcher in \emph{Prof. Tamim Asfour's} robotics lab, where I also met and befriended my close friend \emph{Jonas Rothfuss} while collaborating on robotic perception and manipulation. Tamim further enabled me to join \emph{Prof. Jeannette Bohg's} robotics lab at Stanford, where I had the privilege of meeting and working alongside \emph{Krishnan Srinivasan}, \emph{Michelle Lee}, \emph{Margot Vulliez}, \emph{Peter Zachares}, \emph{Varun Nambiar}, \emph{Lin Shao}, and \emph{Suraj Nair}. I am forever grateful for the opportunities I was given and for these good-hearted people I met during this formative stage of my research journey.\\

Continuing to be fascinated by empirical and automated deep learning research, I joined \emph{Prof. Frank Hutter's} ML/ AutoML Lab in Freiburg. Frank is not only a stellar researcher but also an exceptional supervisor who sparked my interest in meta-learning, gave me the absolute freedom to work independently, facilitated collaborations, and enabled me to explore exciting ideas outside of my thesis scope. As a supervisor, he believes in inspiring and supporting his team rather than imposing ideas which is an approach that has proven highly effective and fosters a happy and collaborative team environment. Beyond his role as a supervisor, he is a great, humble, and social human being. The fact that everyone I meet speaks so fondly of him is a testament to the positive impact he has on those around him. All these traits combined make him a truly unique and rare individual in ML research, and I am deeply thankful for his guidance and for the opportunity to have been able to work closely with him.\\

I also thank my current and former lab colleagues \emph{Arb\"er, André, Arlind, Sam, Sebastian, Eddie, Katharina, Matthias, J\"org, Matilde, Noor, Raghu, Steven, Stefan, Lennart, Thomas, Johannes, Arjun, Rhea, Heri, Ivo, Danny, Max, Jake, Neeratyoy, Julien, David, Timur, Mahmoud, Noah, Robin, Yash, and Frederic} for the shared experiences and conversations over coffee and at conferences, team retreats and hikes, joint team lunches, and board game nights. I thank the secretaries \emph{Svenja, Morgan, Lina, and Petra} for helping me with the bureaucracy.\\

Special thanks to \emph{Arb\"er Zela}, \emph{Raghu Rajan}, \emph{Samuel M\"uller}, \emph{Arlind Kadra}, and \emph{J\"org Franke} for all the long walks around the office and deep talks. I am particularly grateful to have found a great friend in \emph{Arb\"er}, and I genuinely hope that our friendship extends far beyond the PhD journey. I am also grateful for the tennis matches and swims in the Dreisam River with \emph{Raghu, Sam, and Yash Mehta}.\\

Thanks to all the students I have been grateful to have worked with: \emph{Thomas Nierhoff, Ivo Rapant, Diane Wagner, Ekrem \"Ozt\"urk, Moreno Schlageter, Andreas S\"alinger, Muhammad Ali, Dipti Sengupta, Pawel Bugyi, Kashan Karimudin, Maciej Janowski, Albanot Makolli, and Jeta Bekteshi}. I also want to thank all the people with whom I collaborated and published papers.\\

Special mention to my dear friends \emph{Jonas Rothfuss, David Betghe, Colin Weitmann, Steffen Huber, Makai Chapman, Daniel Ziegerer, Krishnan Srinivasan, Erica Gawley, James Gawley, Michael Wall, Heinrich Dinkel, Janika Schmidt, Eleni Triantafillou}, and to my Karlsruhe-crew consisting of \emph{Paulina Matuszak, Michael Kuzmin, Xenia Hoffmeister, Helen Meier, and Julia Specht}. I am also deeply grateful to my former partner \emph{Atessa Schilli}, who accompanied my research journey from the very beginning and supported me unconditionally during a significant part of it.\\

Most importantly, I want to thank my parents, \emph{Elisabeth} and \emph{Florindo}, as well as my sister, \emph{Liana}, my niece, \emph{Amália}, and the rest of my family living in Germany and Portugal for always being there for me: Avó Madalena, Avô Raul, Claudia, Duarte, Valerio, Ausenda, Alberto, and Isabel. Your support, love, and encouragement have been my foundation throughout this journey. By always believing in me, my parents give me the strength to pursue my goals, even in the toughest moments. While my sister and I may have had our quarrels, I learned a great deal about myself through our interactions. My niece's joyful energy has brought light to even the most challenging days. I am forever thankful for your strength, stability, and the role you all played in making this achievement possible.\\

As a passionate listener of electronic music, I am very grateful to producers and labels such as \emph{CRi}, \emph{Keinemusik}, \emph{Afterlife}, and \emph{Above \& Beyond} for crafting immersive tracks that have been my constant companions, fueling creativity and concentration during countless hours of coding and learning. I also thank the invaluable online resources \emph{Google Scholar} and \emph{arXiv} and acknowledge the use of \emph{OpenAI's} ChatGPT as a supplementary tool to assist with brainstorming ideas, suggesting alternative phrasing, refining individual sentences, and generating ideas for structuring the dissertation. I authored and reviewed all final content and decisions to ensure originality and academic rigor. 


\setcounter{tocdepth}{1}
\tableofcontents*

\mainmatter
\part{Introduction}\label{part:intro}
\chapter{Motivation}
The field of \emph{Artificial Intelligence (AI)} has achieved transformational progress in various domains over the past years. From Bioinformatics with three-dimensional protein structure prediction \citep{abramson2024accurate}, capabilities to segment novel objects in videos in real-time \citep{ravi2024sam}, text-to-image-models \citep{stablediffusion3}, to foundation models such as \emph{Large Language Models (LLMs)} \citep{llama3, mixtral, brown-neurips20a, radford-openaiblog19a}, and large multimodal models \citep{flamingo, team2023gemini, gpt4, whisper, radford-icml21a}. While lacking behind in some areas, such as mathematical reasoning, the scaling of models to billions of parameters has allowed these models to surpass human performance on several benchmarks, including visual reasoning and English understanding, with a quickly growing number of open-source models becoming available \cite{kaplan-arxiv20a, AIIndex2024, guo2025deepseek}. However, the availability of models comes with a multitude of risks and challenges, such as privacy concerns, explainability, safety and security, or inference in low-resource environments.\\

A significant challenge that arises from these remarkable advancements and the resulting model abundance is the difficulty of choosing suitable AI systems and finetuning them in practical applications. Not only in the rapidly evolving field of LLMs, but also in fields such as image classification, state-of-the-art pretrained models with incremental improvements are released at a high frequency. For example, at the time of writing, the popular model hub Hugging Face hosts over 500,000 pretrained models, including 340,000 in language processing, $\sim$78,000 in computer vision, and $\sim$14,500 models specifically for image classification ~\cite{huggingface_model_hub}. This surge in the number of models exemplifies the \emph{Jevons Paradox} \citep{jevons_paradox}, which suggests that increased efficiency lowers costs and, paradoxically, results in even higher demand. In the case of LLMs, this paradox is evident in examples like DeepSeek's R1-Zero model \citep{guo2025deepseek}, an open-source release that achieves on-par performance with open and closed-source alternatives while being trained at a fraction of their cost. As training LLMs becomes cheaper and training pipelines more effective, this trend will likely accelerate further, driving the proliferation of available models.\\

\begin{figure}[t]
    \centering
    \begin{minipage}[t]{0.48\textwidth}
        \centering
        \includegraphics[height=0.53\textwidth]{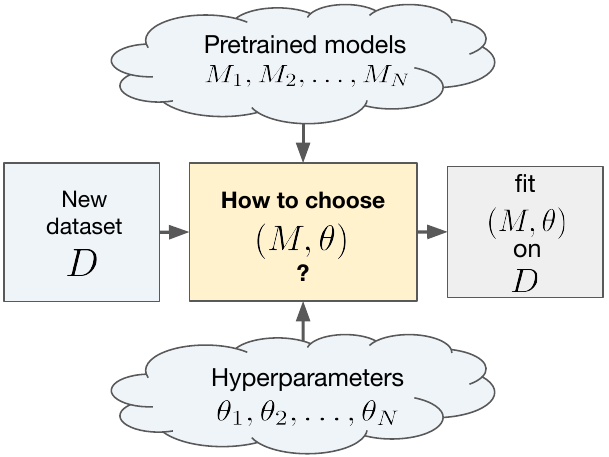}
        \caption*{(a)}
        \label{fig:cash}
    \end{minipage}
    \hfill
    \begin{minipage}[t]{0.48\textwidth}
        \centering
        \includegraphics[height=0.53\textwidth]{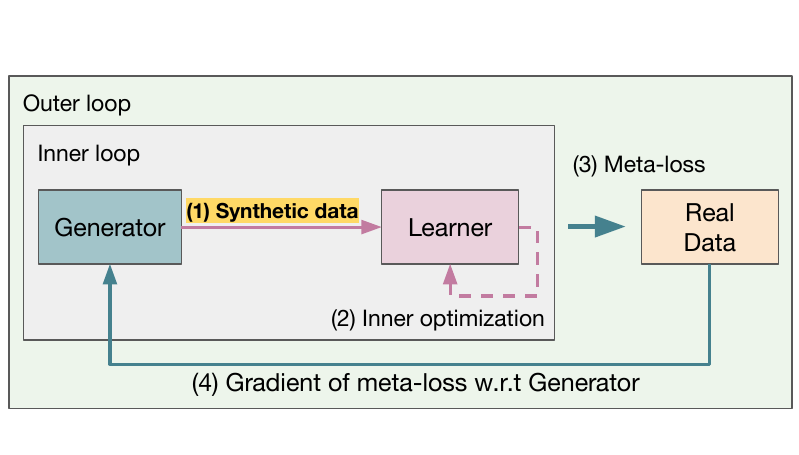}
        \caption*{(b)}
        \label{fig:synthetic_data}
    \end{minipage}
    \caption{(a) The Combined Algorithm and Hyperparameter Optimization (CASH) problem visualized. (b) An exemplary meta-learning framework for learning synthetic data generators (visualization inspired by \citet{such-icml20a}).}
    \label{fig:cash_synthetic_data}
\end{figure}

\begin{wrapfigure}{r}{0.4\textwidth}
    \centering
    \vskip -.15in
    \includegraphics[width=0.4\textwidth]{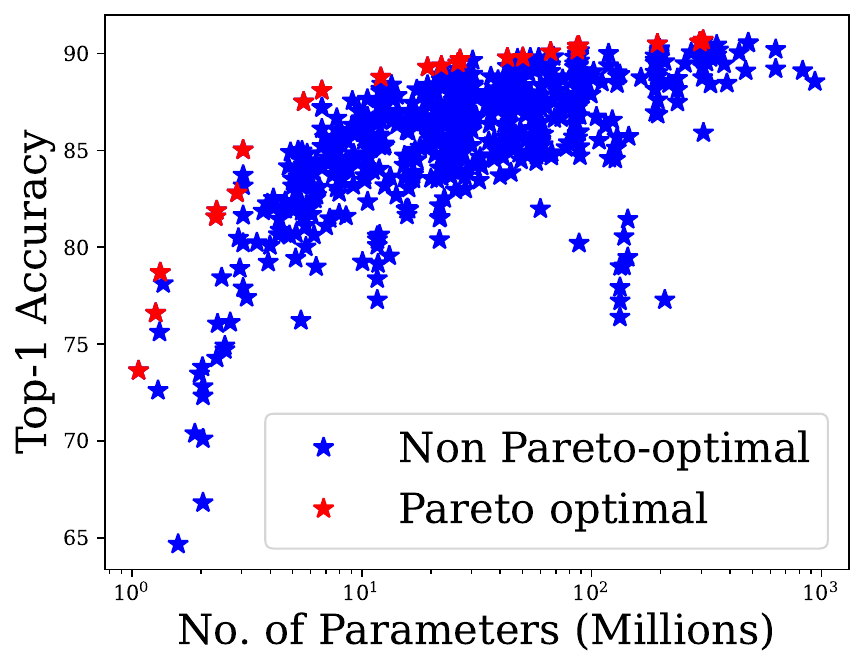}
    \caption{Illustrative example of the large set of over 700 pretrained image classification models available from the timm library, with still 24 models on the Pareto front (illustration from the Quick-Tune paper \cite{pineda-iclr24a}).}
    \vspace{-5pt}
    \label{fig:timm_models}
\end{wrapfigure}

As illustrated in \Cref{fig:cash_synthetic_data} (left), this abundance presents practitioners with a complex question: how to select the most suitable \emph{deep learning (DL) pipeline} \citep{rumelhart-book85a, lecun-nc89a, krizhevsky-nips12a, goodfellow-mit16a}, which consists of a pretrained model and its finetuning hyperparameters, for a specific new dataset or task? Considering a practical example, \Cref{fig:timm_models} visualizes this challenge from a practitioner's standpoint. It shows 700 pretrained image classification models taken from the popular timm library \cite{rw2019timm}, plotted by accuracy against model size (measured in the number of parameters). Even when restricting the choice to models on the Pareto optimal front, practitioners must still select from a subset of 24 models. Additionally considering other fidelities, such as inference time or memory constraints, may further drive the complexity of selecting a DL pipeline to the specific needs of their applications. Lastly, this problem is almost universal: not only in computer vision we see an explosion of pretrained model availability but a similar development can be already observed in the language model domain, where many institutions are competing for the best LLM for coding, conversation, or reasoning.\\

However, with the recent developments in model scaling, we argue that selecting the best pipeline alone is insufficient without leveraging data effectively. For instance, \citet{hoffmann2022training} has shown that compute-optimal training strategies require scaling the model parameters proportionally with the dataset size to fully realize their potential. However, achieving this assumes often high-quality human data, which is often expensive to obtain or entirely unavailable for certain domains. Meanwhile, the AI Index Report 2024 \cite{AIIndex2024} indicates that the availability of high-quality language data may soon be depleted, raising concerns about the long-term sustainability of model scaling given this bottleneck.\footnote{According to \citet{AIIndex2024LackOfData}, high-quality language data is depleted by 2024, and low-quality language data within two decades.} 
Recently released LLMs such as Llama 3.3 already indicate the increasing reliance on synthetic data, using up to 25M synthetic samples to augment the finetuning corpus \citep{llama-3-3-modelcard}. We believe this motivates developing strategies to maximize the effectiveness of existing data, such as enhanced data augmentation methods, optimizing existing augmentation strategies, and generating synthetic data to complement real-world data. These challenges emphasize the dual need to navigate the abundance of pretrained models and their finetuning hyperparameters, as well as the effective use of data.

\paragraph{Navigating the Ocean of Deep Learning Pipelines}
Choosing an appropriate DL pipeline for a new task can be compared to navigating an ocean. Imagine a navigator that needs to choose the best ship (pretrained model) and adjust its sails (hyperparameters) to reach a new destination (dataset or task to finetune to). Each ship has its own capabilities, size, and speed. Akin to a practitioner, the navigator must not only select an appropriate ship for their journey but also choose an overwhelming number of settings for arriving at the destination, such as the crew size, cargo capacity, rudder, or sail orientation.\footnote{for simplicity, this illustration disregards corrections of chosen settings during the journey} Practitioners have to rely on heuristics and intuition or trial-and-error strategies if they have no prior experience. They risk choosing an ill-suited ship, and misselections of such settings may result in the navigator not reaching the destination or reaching it weeks later. While this illustration may seem far-fetched, it emphasizes how challenging the task of making the right decision under uncertainty and a large number of available options. Analogously, choosing an ill-suited DL pipeline can lead to many different problems, such as instability in training, incomplete convergence, under and overfitting of the task and many more. This is aggravated significantly if a practitioner has little or no prior knowledge to make informed decisions. Consequently, it makes sense to find automated and algorithmic solutions to help address this problem and outsource the need for such skills and knowledge to an algorithm.\\

In Machine Learning (ML), already in the 1970s, this problem was formulated as a search problem by Rice, who introduced the \emph{Algorithm Selection} problem \cite{rice-aic76a}. Building on that, the \emph{Combined Algorithm Selection and Hyperparameter Optimization problem (CASH)}~\citep{thornton-kdd13a, hutter-book19a} was introduced, which treats this search as a joint optimization problem in the space of both model (algorithm) and hyperparameters (see \Cref{fig:cash_synthetic_data} (right)). This problem can be tackled from various angles, for instance, regression~\citep{xu-jair08a}, classification~\citep{xu-sat12a}, or clustering-based~\citep{kadioglu-ecai10a}. These approaches rely on \emph{meta-learning} \citep{schmidhuber-tum87a, andrychowicz-neurips16a, finn-icml17a, vanschoren-automlbook19a, hospedales-tpami21a}, i.e. learning from prior experience, to train a \emph{selector} using prior recorded performance data about algorithms on similar problem instances (e.g., datasets), combined with their characteristics, to map new problems to the best algorithm and hyperparameters. Depending on how well the prior data on which the selector was trained resembles future problem instances, the selector is equipped to make well-grounded decisions about which algorithm to apply to new problem instances. To go back to the example, this methodology aims to replace the navigator with one that is capable of meta-learning, i.e. a navigator that exploits past experiences of navigating various ships to different destinations, enabling informed decisions for new journeys. In ML, this approach has mostly been applied to classic ML models that are less complex (e.g., fewer hyperparameters) and less computationally expensive to train compared to Deep Learning models. However, adapting this methodology to DL pipelines introduces new challenges due to the scale and complexity of pretrained models and their hyperparameters. This dissertation aims to bridge this gap by extending the CASH framework to Deep Learning by leveraging meta-learning.

\paragraph{What about the data?}
Above, we touched on the duality between selecting a DL pipeline and the effective use of data. However, ensuring the effective use of data in machine learning presents several challenges. Returning to the running example, a successful journey depends a lot on the availability of good ships. For instance, the navigator might reinforce parts of a ship to develop specialized ships tailored to different types of journeys: an all-rounder ship capable of handling a variety of conditions or a specialist ship optimized for a specific type of journey (e.g. navigating narrow canals). Analogously, in machine learning, one way to achieve effective use of data is to generate or adapt datasets that are appropriate for the task at hand. This can be attained through data augmentation \citep{chawla-jair02a, simard-icdar03a, devries-arxiv17a, image_augmentation_survey} or synthetic data generation \citep{jakobi1995noise, goodfellow-nips14a, kingma-iclr14a, tobin2017domain, sankaranarayanan-cvpr18a, tremblay2018training, ho-neurips20a}, either individually or by mixing them with real data to diversify and expand datasets. These approaches can be an important tool for enhancing robustness (e.g., as measured by a low variance in predictions) and task-specific performance in both pretraining and finetuning DL models \citep{image_augmentation_survey, cubuk-cvpr19a, grill-neurips20a, such-icml20a, tobin2017domain}. Despite their potential, data augmentation and synthetic data generation are often treated heuristically, relying on manual adjustments that are time-consuming and may be suboptimal. For example, such strategies remain underexplored in typical pretraining regimes like \emph{Self-Supervised Learning (SSL)} \citep{grill-neurips20a}.

To address this challenge, meta-learning offers a promising solution. For instance, data augmentation can be seen as an input for a meta-learner to help it leverage diverse augmented datasets and inform its decisions. Just like a navigator may leverage prior knowledge from past journeys, meta-learning can use prior experience to inform the development of effective data augmentation strategies. For example, one can think of applying data augmentation on a dataset level to augment the data as input to train a previously described (meta-)selector by varying characteristics of the data and help prepare the selector for rare or unseen scenarios. 

It is also viable to treat data augmentation as an optimization hyperparameter since practitioners also often rely on trial and error to identify good configurations. Consequently, incorporating data augmentation as a hyperparameter allows systematic optimization of it, reducing reliance on manual heuristics while improving model robustness. Finally, just as the navigator might innovate entirely new ships by simulating novel scenarios or anticipating challenges, meta-learning can be used to generate data augmentations or synthetic data as an addition or replacement to existing data and tasks. In such cases, synthetic data can be meta-learned and used as training data for subsequent or downstream tasks. This approach is exemplified in \Cref{fig:cash_synthetic_data} (right), which illustrates a general meta-learning framework for optimizing synthetic data generators. The framework uses the performance of a learner trained on real data to optimize the synthetic data generator. For instance, this dissertation proposes this framework for Reinforcement Learning and demonstrates how synthetic environments can be learned for efficient subsequent agent training. Synthetic data generation is especially useful when real-world data collection is expensive (e.g., in robotics). Moreover, synthetic data provides a chance to embed desirable properties, such as enhancing learning efficiency, compensating for underrepresented classes, or introducing controlled variability to better prepare models for real-world complexities.

This dissertation explores the use of meta-learning to develop advanced data augmentation strategies and learn synthetic data generators with the goal of enhancing data effectiveness and efficiency during training while also reducing practitioners' reliance on heuristics and trial-and-error. In this dissertation, these efforts are conducted for the automation of DL pipeline and pretraining in SSL with applications in computer vision but also extend to model-based Reinforcement Learning.

\section{In a Nutshell}
\label{sec:nutshell}
This dissertation explores meta-learning to reduce reliance on manual tuning, trial-and-error, and the use of heuristics in Deep Learning by developing solutions for automated pretraining and finetuning. It extends the concepts of \emph{Automated Machine Learning (AutoML)} \citep{hutter-book19a} to Deep Learning and uses meta-learning to enhance data augmentation methods and generate synthetic data for effective and efficient optimization in Deep Learning. Together, these contributions allow us to address the overarching question of the dissertation:

\begin{tcolorbox}[
    colback=blue!5,    
    sharp corners,     
    boxrule=1pt,       
    width=\textwidth,  
    fonttitle=\bfseries 
]
\emph{How can meta-learning and synthetic data advance automated pretraining and finetuning?}
\end{tcolorbox}

For the purposes of this dissertation, \emph{synthetic data} collectively refers to both data augmentation and synthetic generation methods since both seek to artificially expand or alter the data distribution. To emphasize their differences, data augmentation, and synthetic data generation will be distinctly identified when specifically discussed.

In particular, this dissertation includes contributions that individually concern pretraining, finetuning, data augmentation, and synthetic data generation, as well as works that explore the intersection of these themes to advance automation in Deep Learning pipelines.

\section{Organization of the Dissertation}
This dissertation is a \emph{cumulative dissertation}. As such, it consolidates contributions from multiple research publications by the author. Moreover, all publications listed in this dissertation aim to answer an overarching scientific question described in \Cref{sec:nutshell}. The dissertation is organized into four parts, arranged by topic rather than chronological order:
\begin{itemize}
    \item \textbf{\BFullref{part:intro}}: the current part, introduces the overarching theme of the dissertation, outlines its primary goals, and provides the relevant background context.
    \item \textbf{\BFullref{part:meta-learning-cash}}: contains the set of publications that cluster contributions on meta-learning for the selection and finetuning of DL pipelines. It encompasses works in the domains of image classification and large language models.
    \item \textbf{\BFullref{part:meta-learning-synthetic-data}}: summarizes the second set of publications, which contains contributions focussed on meta-learning data augmentation and synthetic data for better learning. 
    \item \textbf{\BFullref{part:conclusion}}: concludes the dissertation with a summarization of the key takeaways, highlights the achievements of this work, and describes potential avenues for future research in the domain of ML and AI.
\end{itemize}

The background given in Part \ref{part:intro} is recommended reading if the reader is unfamiliar with meta-learning. Moreover, the publications provided in Parts \ref{part:meta-learning-cash} and \ref{part:meta-learning-synthetic-data} are self-contained and can be read independently. The primary distinction between these two parts lies in their focus: Part \ref{part:meta-learning-cash} centers on automating model selection and finetuning in DL pipelines, with data augmentation and synthetic data functioning as secondary components, while Part \ref{part:meta-learning-synthetic-data} addresses directly optimizing data augmentation and synthetic data generation utilizing meta-learning techniques or supporting them to enhance learning. Lastly, the dissertation operates in the following machine learning areas: self-supervised learning, supervised learning, model-based reinforcement learning, in-context learning, automated machine learning, image classification, question-answering, and large language models.

\chapter{Goals of this Dissertation}
As the field of AI grows in scale and complexity, the deployment and finetuning of models become increasingly harder. Practitioners are challenged with the question of how to select and finetune pretrained models for new tasks and how to leverage data effectively to improve training outcomes. This challenge is compounded by the reliance on manual heuristics and the depletion of training data. Addressing these issues requires innovative strategies to automate pipeline selection and finetuning, as well as optimizing data usage. Guided by these challenges, this dissertation seeks to answer the overarching scientific question: \emph{``How can meta-learning and synthetic data advance automated pretraining and finetuning?"} To achieve this, we derived the following research questions that guided our contributions.

\section{Key Challenges}

\subsection*{Challenge 1: Automating Model Selection and Finetuning Amidst a Growing Landscape of Pretrained Models}\addcontentsline{toc}{subsection}{Challenge 1: Automating Model Selection and Finetuning Amidst a Growing Landscape of Pretrained Models}
With the rapid increase in available pretrained models across various domains, identifying an optimal pretrained model and its hyperparameters for finetuning to a new dataset has become increasingly challenging. This task is known as the \emph{Combined Algorithm Selection and Hyperparameter Optimization (CASH)}~\cite{thornton-kdd13a} problem. To illustrate this increase, Hugging Face now hosts over 500,000 pretrained models (of which, for instance, $\sim$78,000 are computer vision and $\sim$340,000 are language processing models)~\cite{huggingface_model_hub}, making manual selection impractical. Moreover, finetuning involves numerous hyperparameters, and optimizing them is not only time-consuming but often infeasible for practitioners who have to rely on heuristics. Existing meta-learning-based efforts to automate model selection and hyperparameter tuning have primarily focused on traditional Machine Learning (ML) models~\cite{feurer-automlbook19b, kotthoff-jmlr17a}, do not leverage dataset information, and usually do not consider model selection and hyperparameter optimization as a joint problem. A key challenge lies in adapting these CASH techniques, traditionally used in non-Deep Learning settings, to the vastly more computationally expensive Deep Learning landscape and in effectively using meta-learning to exploit prior training data to execute automated finetuning on new tasks. This includes finding solutions that are scalable and generalizable across domains, such as extending techniques that cover automating finetuning of large language models or image classification. 



\subsection*{Challenge 2: Enhancing Learning With Advanced Data Augmentation and Synthetic Data Generation}\addcontentsline{toc}{subsection}{Challenge 2: Enhancing Learning With Advanced Data Augmentation and Synthetic Data Generation}

Building upon the challenge of selecting and finetuning pretrained models in a growing landscape, a central assumption is to develop robust, powerful, and general pretrained models. As models grow larger and much of the available human-generated data is already being used for training, the demand for extensive training datasets continues to increase. In Self-Supervised Learning (SSL), for example, a common approach to extend the pretraining dataset is to rely heavily on data augmentation, making the data augmentation strategy and its hyperparameter settings essential for downstream performance~\cite{chen-icml20b, grill-neurips20a}. However, in SSL, data augmentation remains underexplored due to instability at high augmentation intensities and the cost of identifying effective techniques. The challenge of adopting advanced data augmentation also extends to automated model selection and hyperparameter tuning (Challenge 1), where dataset level augmentation can enable meta-learned model selectors to cope better with variance encountered in datasets at test time. Yet, best practices on how to use data more effectively in these domains are scarce within the scientific community and remain to be explored further. An alternative to data augmentation lies in synthetic data generation. Unlike augmentation, synthetic data can be generated and used more independently of real data, for instance, by sampling synthetic data from distribution priors or by directly (meta-)learning the data. Synthetic data is particularly valuable in domains where collecting real-world data is costly, such as in Reinforcement Learning (RL). For example, in robotics, this is further amplified by the trend toward developing generalist models trained across multiple robot platforms and datasets~\cite{chen-generalizable-robotic-reward, physical_intelligence_pi0}. Consequently, the need for synthetic datasets and rich simulations allowing minimal real-world interactions is growing. However, creating synthetic data that captures real-world characteristics while balancing diversity and training efficiency remains a significant hurdle~\cite{AIIndex2024LackOfData}. Therefore, we believe a core challenge lies in exploring more ways to leverage data effectively by advancing data augmentation and synthetic data generation.
While meta-learning offers a promising avenue for leveraging prior knowledge to inform advanced data augmentation and synthetic data generation, its potential remains underexplored.

\subsection*{Challenge 3: Ensuring Reproducibility and Practical Applicability of Automated Learning Methods}\addcontentsline{toc}{subsection}{Challenge 3: Ensuring Reproducibility and Practical Applicability of Automated Learning Methods}

Another challenge we face in machine learning research is ensuring reproducibility and practical applicability~\cite{pineau-reproducibility-in-ml}. Research artifacts are often not made open-source, and datasets or model checkpoints are not released, which results in limited accessibility. Even when research artifacts are made publicly available, they often remain at the research level due to weak incentives for creating reproducible research. Consequently, research methods are challenging to implement without guidance, and high computational demands can pose an additional barrier to replication and adoption. Enabling reproducibility and adopting research findings within and beyond the machine learning community is challenging. It requires more than making code and data available. Lowering the barrier to entry also requires tutorials, tools with standardized interfaces to allow for application across domains, benchmarks for feature comparison, and more. 


\section{Contributions}
To address the key challenges identified in the previous section, we formulate a set of research questions that shape the contributions of this cumulative dissertation. We then provide an abstract overview of how the dissertation as a whole addresses these questions. Since each work within this dissertation is presented as an individual chapter, we also provide individual summaries of those along with how they address the overarching scientific question at the end of this section.

\subsection*{Research Questions}
Based on the challenges outlined above, this dissertation seeks to answer the following research questions (RQ) and describes how they address these key challenges:

\begin{itemize}
    \item \textbf{Challenge 1: Automating Model Selection and Finetuning Admidst a Growing Landscape of Pretrained Models}
    \begin{itemize}
        \item \textbf{RQ 1:} Given a set of pretrained models, how can one meta-learn to automatically select the best pretrained model and its hyperparameters to finetune it to a new dataset? 
        \item \textbf{RQ 2:} To what extent can these methods generalize to unseen datasets?
        \item \textbf{RQ 3:} How do these methods generalize across different learning domains such as vision and language? 
    \end{itemize}

    \item \textbf{Challenge 2: Enhancing Learning Through Advanced Data Augmentation and Synthetic Data Generation}
    \begin{itemize}
        \item \textbf{RQ 4:} How does data augmentation impact the effectiveness of model pretraining in Self-Supervised Learning (SSL), and how can meta-learning be leveraged to optimize augmentation strategies?
        \item \textbf{RQ 5:} How can meta-learning be leveraged to generate synthetic environment models that act as a replacement to Reinforcement Learning (RL) environments and that improve agent training in RL? 
        \item \textbf{RQ 6:} What potential does randomly sampled synthetic data hold for optimizing synthetic environment models, and to what extent can these models be utilized for agent training in RL?
    \end{itemize}

    \item \textbf{Challenge 3: Ensuring Reproducibility and Practical Applicability of Automated Learning Methods}
    \begin{itemize}
        \item \textbf{RQ 7:} What practices and tools can be implemented to ensure that the developed automated learning methods are reproducible, accessible, and practically applicable to the broader machine learning community? 
    \end{itemize}
\end{itemize}

We address research questions 1-3 as well as question 7 in \Cref{part:meta-learning-cash}. In particular, we address the first two questions by proposing two model-based and meta-learned methods, \emph{ZAP} and \emph{Quick-Tune}, which provide solutions for supervised learning in computer vision and are detailed in \Cref{chap:zap,chap:quicktune}. Both methods also address the \emph{second research question} by contributing to generalization across unseen datasets in empirical experiments. While ZAP incorporates dataset level augmentation to enhance its meta-dataset, Quick-Tune further optimizes augmentation strategies and hyperparameters. Addressing the \emph{third research question}, we show the generalization of automated model selection and finetuning to other learning domains through the aid of \emph{Quick-Tune-Tool}, a tool derived from Quick-Tune with a standardized interface. We use it to adapt Quick-Tune to the domain of Large Language Models, which we discuss in \Cref{chap:qtt,chap:qt_llms}. The \emph{seventh research question}, which revolves around practices and tools to enable access for the broader machine learning community, is addressed primarily by efforts made in Quick-Tune-Tool (\Cref{chap:qtt}) and by the code releases for all papers in this dissertation except for the work given in \Cref{chap:qt_llms} which is the only work without a code release.

With \Cref{part:meta-learning-synthetic-data}, we turn to data augmentation and synthetic data and address research questions 4-6. We provide empirical insights to the \emph{fourth research question} about the effect of data augmentation in SSL-based pretraining of vision models in \Cref{chap:on_the_importance_ssl}. Based on these findings, we introduce \emph{Hard View Pretraining (HVP)} in \Cref{chap:hvp}, where we devise an advanced data augmentation strategy that uses single-task meta-learning.
Addressing the \emph{fifth research question}, \Cref{chap:learningses} introduces \emph{Synthetic Environments}, meta-learned copies of real RL environments that allow efficient agent training through their synthetic dynamics and rewards. In \Cref{chap:oswm}, we explore using randomly sampled synthetic data for training a \emph{One-Shot World Model}, a generalist synthetic environment and hereby address the \emph{sixth question}.\\

\subsection*{Summary of Contributions}
Each chapter of this cumulative dissertation corresponds to a previously peer-reviewed and published work. We now provide a more detailed summary of each work. We explain how these works address the individual research questions and contribute to the overarching scientific question of this dissertation: \emph{``How can meta-learning and synthetic data advance automated pretraining and finetuning?"}

\subsubsection*{\Fullref{chap:zap}}
This chapter presents a solution to the central problem of this dissertation: Given a new dataset $D$, how can a pretrained model be chosen to finetune on $D$ and set its hyperparameters? In this work, we define the combination of a pretrained model and the finetuning hyperparameters as {Deep Learning (DL) Pipeline. Our approach, \emph{Zero-shot AutoML with Pretrained Models (ZAP)}, addresses this as a Combined Algorithm Selection and Hyperparameter Optimization (CASH) problem. ZAP involves creating a meta-dataset with performances of various DL pipelines on a broad range of image classification datasets. We also apply data augmentation at the dataset level, for instance, by varying the number of classes or samples to increase the size of the meta-dataset. Using this meta-dataset, we meta-learn a surrogate model that selects the best DL pipeline given meta-features (e.g., image resolution) in a zero-shot setting, i.e., without exploratory evaluations of the target dataset.

Inspired by \emph{Algorithm Selection (AS)} techniques, our zero-shot model selection approach optimizes a ranking objective across datasets, selecting DL pipelines based on both pipeline hyperparameters and dataset meta-features. A key contribution is to frame DL pipeline selection as a classical AS problem and extend it by modeling pipelines as points in a latent space, enabling performance insights from some pipelines to inform predictions for others. To the best of our knowledge, we are the first to adopt AS methods with meta-features beyond classical machine learning algorithms and apply them to pretrained DL models for image classification. 

Consequently, we propose two approaches: one tackling ZAP with AS methods and one tackling it by using a neural network selector optimized with a ranking objective over DL pipelines.  These approaches were evaluated in the 2019 ChaLearn Automated Deep Learning Challenge, outperforming other participants under budget constraints and anytime performance metrics.  Furthermore, we conduct studies on selector performance when trained with a sparse meta-dataset. 

\paragraph{How does this work address the overarching scientific question?}
ZAP addresses the dissertation's central question by introducing a zero-shot, meta-learned surrogate selection model to automate model selection and hyperparameter tuning for finetuning to image classification datasets. ZAP demonstrates that meta-learned insights from diverse datasets can inform zero-shot model selection and hyperparameter tuning for new datasets. Additionally, it shows that dataset level augmentation effectively enhances selection performance. This aligns with the dissertation's goals by illustrating how meta-learning, amplified by synthetic data or data augmentation, enables efficient and adaptive automation of model selection and finetuning in computer vision. Lastly, its meta-dataset-based approach is generalizable, extending applicability to other learning domains as well.

\subsubsection*{\Fullref{chap:quicktune}}
In this chapter, we present \emph{Quick-Tune}. Similar to ZAP, Quick-Tune also addresses the problem of automated model selection and finetuning given an unseen dataset (represented by meta-features) and a set of pretrained models (model hub) through the lenses of Combined Algorithm Selection and Hyperparameter Optimization (CASH) and image classification. As such, Quick-Tune builds on ZAP’s foundation but takes a \emph{few-shot}, Bayesian framework-based, and cost-adaptive approach. In contrast to ZAP, which operates in a zero-shot setting (i.e., without updating any meta-model(s)) and uses final performance scores for model selection, Quick-Tune uses partial learning curves as few-shot observations to meta-learn and update its probabilistic predictor (surrogate model) during training, allowing it to predict the performance of pipelines. Additionally, Quick-Tune meta-learns a second model, a cost estimator that estimates the runtime cost of a pipeline. This makes Quick-Tune a multi-fidelity approach, allowing users to specify time budget constraints and adjust the meta-search time to practical limitations. Moreover, Quick-Tune also integrates various finetuning and augmentation strategies into its pipeline search space. Both models, performance predictor and cost estimator, are then used within a Bayesian Optimization framework to form an Expected Improvement-grounded acquisition function, allowing the selection of the next pipeline to maximize expected performance in a cost-sensitive manner. In contrast to ZAP, which approaches the CASH problem in a \emph{zero-shot} way, i.e., without requiring observations from the target dataset, Quick-Tune tackles it with a \emph{few-shot} way. It explores its search space of pipelines by iteratively refitting its performance and cost model based on self-collected observations (shots) from pipeline evaluations during search. All in all, we show that Quick-Tune can outperform both hyperparameter optimization tuning methods and finetuned single large vision models across a large collection of image classification datasets.

\textbf{How does this work address the overarching scientific question?}
Quick-Tune, like ZAP, offers an adaptive solution to the central question of how to automatically select and tune pretrained models for image classification using meta-learning. While both leverage meta-learning, Quick-Tune differs by applying a few-shot, Bayesian gray-box optimization based on partial learning curves.  This few-shot approach allows not only iterative refinement but also to adhere to time or budget constraints set by practitioners. We also provide empirical evidence that using Quick-Tune can be superior to learning classification heads in combination with large-scale and state-of-the-art feature extractor backbones. Different from ZAP, Quick-Tune does not apply data augmentation on the dataset level but incorporates data augmentation hyperparameters such as the augmentation strategy, operations, and magnitude into its pipeline search space. 


\subsubsection*{\Fullref{chap:qtt}}
With \emph{Quick-Tune-Tool (QTT)}, as the name suggests, we aim to make the Quick-Tune algorithm more widely accessible. The main goal of this tool is to provide a standardized interface to facilitate adoption beyond the image classification to domains like image segmentation or large language models (see \Cref{chap:qt_llms}). QTT further lowers the entry barrier for adopters by providing an architectural overview, a guideline for running Quick-Tune for image classification, and additional example experiments. The tool also supports models from major model hubs like Hugging Face and timm.

\textbf{How does this work address the overarching scientific question?}
Open-sourcing research code often falls short of enabling a wide adoption in and beyond the research community. Automating pretraining and finetuning with meta-learning can be complex as it requires leveraging prior training and validation data. By releasing QTT as a user-friendly tool, we aim to advance meta-learning research by making it more efficient and allowing advancements in other domains. Additionally, its interface is designed to accommodate data augmentation in the model search space, further enhancing its flexibility.

\subsubsection*{\Fullref{chap:qt_llms}}
In this chapter, we present work that uses Quick-Tune-Tool to explore the applicability of the Quick-Tune algorithm to the language processing domain. More precisely, we investigate whether we can use Quick-Tune to learn to transfer configurations for finetuning Large Language Models (LLM). With this exploratory contribution, we demonstrate the adaptability of meta-learning-based automation beyond image classification. In order to train our meta-models in this case, we created a meta-dataset containing learning curves, meta-features, and cost values. We created it by generating synthetic question-answer (QA) datasets from 30 scientific papers by instructing a Llama-3.1-70B model to extract atomic facts. For each paper, we trained 60 random and default finetuning pipelines and evaluated them using a teacher-student framework, where the teacher (LLama-3.1) model assessed the correctness of answers of the student (finetuned LLM), resulting in a total of 1,800 run recordings. Based on this meta-dataset, we pretrain the performance and cost predictors using the original Quick-Tune logic and search for well-performing finetuning configurations with its acquisition function.
Tested on unseen synthetically generated QA datasets, our method outperforms random search, a popular hyperparameter optimization method, and a default finetuning pipeline.

\textbf{How does this work address the overarching scientific question?}
This work addresses the overarching scientific question of applying meta-learning to develop methods for automating model selection and finetuning in a learning domain beyond vision: language processing. Additionally, we show how a tool such as Quick-Tune-Tool facilitates this effective transfer. Aligning with the dissertation's emphasis on leveraging synthetic data, we also demonstrate the effective use of synthetic data by generating question-answer datasets to enable meta-learning.

\subsubsection*{\Fullref{chap:on_the_importance_ssl}}
Coming to \Cref{part:meta-learning-synthetic-data}, we now turn towards using enhanced data augmentation and synthetic data to achieve more effective pretraining and learning itself. In this chapter, we present a work in computer vision illuminating the underestimated role and importance of hyperparameters and data augmentation in Self-Supervised Learning (SSL), a popular methodology used for pretraining.

We use Bayesian optimization for tuning training (e.g., learning rate) and data augmentation (e.g., image distortion magnitude) hyperparameters using the popular SimSiam SSL approach trained and evaluated on CIFAR-10, CIFAR-100, and a medical dataset. Overall, this work illustrates that optimizing training hyperparameters only leads to marginal improvements, while optimizing data augmentation hyperparameters results in consistent performance improvements (1-2.3\%) under the standard linear evaluation protocol. These improvements show empirically that SimSiam's training hyperparameters are already well-tuned (and also other SSL approaches as reported in \ref{chap:hvp}). Additionally, this study also reports on hyperparameter importance, further highlighting the potential for improvement in data augmentation hyperparameters. As a consequence of these observations, the paper introduces an automated data augmentation method, GroupAugment, that operates on groups of augmentations (e.g., color transformations) and designs sampling strategies over these groups. Automated data augmentation methods have predominantly been investigated in the context of supervised or semi-supervised learning, but their application to SSL remains underexplored \citep{cubuk-cvpr19a, sungbin-neurips19a, lim-neurips19a, cubuk-cvpr20a, muller-iccv21a, reed-cvpr21a}. We also report performances of other popular automated data augmentation methods introduced for Supervised Learning when applied to SSL, finding that in the majority of cases, GroupAugment is able to outperform these methods.

\textbf{How does this work address the overarching scientific question?}
While many works acknowledge the critical role of data augmentation in SSL, little work identifies the importance of data augmentation hyperparameters and the potential gains achievable through sophisticated data augmentation strategies. This work uncovers that data augmentation is underexplored in SSL. By showing that improved data augmentation yields consistent benefits, this paper sets the stage for the next chapter, which develops a more effective data augmentation strategy in pursuit of automated pretraining techniques.

\subsubsection*{\Fullref{chap:hvp}}
The work presented in this chapter builds on the findings made in \Cref{chap:on_the_importance_ssl} that empirically uncovered the underestimated role of data augmentation in discriminative SSL. In this work, we leverage these findings and develop a simple but effective data augmentation strategy, \emph{Hard View Pretraining (HVP)}, generally applicable to discriminative SSL methods. HVP is a learning-free augmentation strategy that extends the widely used random view generation. It exposes the model to challenging samples during pretraining by randomly sampling views and selecting adversarially the ones with the highest loss according to the current training progression during each training step. Compared to standard SSL, our approach introduces a challenging learning scenario in which the model is encouraged to learn more discriminative features by being exposed to harder views. At the beginning of training, the embedding space lacks a defined structure for representing similarity among views. With training progression, HVP refines the concept of similarity through exposure to increasingly harder views. While automated data augmentation has been traditionally investigated in the context of supervised learning, HVP demonstrates its potential in self-supervised settings by automating the selection of challenging augmentations that adapt to the model’s state during training. This simple but effective strategy consistently and significantly improves downstream task performance across four popular SSL methods when trained with ResNets and Vision Transformers (ViT) on ImageNet. Underpinning the effectiveness and scalability of HVP, we achieved a new state-of-the-art result on the ViT-B/16 model architecture. We also show empirically that HVP regularizes models to be more robust to hyperparameter variations when used for downstream tasks. 

\textbf{How does this work address the overarching scientific question?}
HVP serves as a first step toward automating pretraining by showcasing that challenging and model-state-dependent data augmentation strategies can consistently yield better downstream task performance. HVP can also be viewed as synthetically generating more diverse and difficult training data by selecting hard views based on their loss, incorporating a notion of difficulty into the augmentation process. Since HVP operates at a meta-level by adversarially selecting hard views based on their loss, it can be interpreted as single-task meta-learning, which directly addresses the overarching scientific question. By exposing models to increasingly challenging scenarios during pretraining, HVP also makes them more robust to hyperparameter variations, thereby simplifying the finetuning process for various vision architectures. This combination of improved augmentation, robustness, and scalability positions HVP as a key advancement toward automated and effective SSL pretraining.

\subsubsection*{\Fullref{chap:learningses}}
This chapter presents a work that bridges the two central themes of this dissertation: meta-learning and synthetic data. More precisely, we propose using meta-learning in the form of bi-level optimization to learn synthetic neural data generators. We introduce and explore this framework in Reinforcement Learning (RL) since it offers a toy-like environment with reasonable computational costs. The goal of this work is to meta-learn \emph{Synthetic Environments (SEs) and Reward Networks (RNs)} that provide synthetic proxies to target RL environments (defined as Markov Decision Processes). SEs mimic both state dynamics and rewards, while RNs focus solely on modeling rewards of the target (or real) environment. Algorithmically, we use two nested loops: in the inner loop, we train RL agents on the proxy. In the outer loop, we evaluate the agents on the target environment and use their performance to optimize the proxy parameters, which, over time, evolve to resemble more performant compressions of the target environment. We evaluate this approach on a broad range of RL algorithms and classic control environments. Empirically, we show that SEs and RNs are not only able to train agents to solve real environments but can also be trained to be more efficient and robust to agent hyperparameter variations compared to real environments. Our results indicate that synthetic proxies achieve this performance by learning informed representations that guide the agents toward relevant states. Lastly, the SEs are not only robust to hyperparameters but can also transfer to train unseen agent algorithms.

\textbf{How does this work address the overarching scientific question?}
In the context of leveraging meta-learning and synthetic data for automated pretraining and finetuning, the works presented in this dissertation so far have primarily used synthetic data and data augmentation as tools to enhance pretraining and finetuning rather than learning them directly. While HVP (Chapter \ref{chap:hvp}) represents a step toward automated data augmentation for pretraining through single-task meta-learning, it does not explicitly optimize the data augmentation itself. Similarly, GroupAugment (Chapter \ref{chap:on_the_importance_ssl}) combines existing data augmentation methods with hyperparameter optimization to improve pretraining but does not directly learn augmentations. In contrast, the framework of Synthetic Environments (SEs) takes a step forward by directly learning synthetic data through meta-learning in the context of RL. The emergent properties of SEs and RNs, such as more efficient training, hyperparameter robustness, and agent-agnostic adaptability, contribute to advancing automated and efficient optimization of RL agents. This not only addresses the dissertation's goal of automated learning but also aligns with the broader trend of developing generalist RL models.



\subsubsection*{\Fullref{chap:oswm}}
In the previous chapter, we introduced a meta-learning framework that directly learns synthetic compressions (SEs and RNs) as proxies for specific RL environments by optimizing them based on agent performance. In this chapter, we propose the \emph{One-Shot World Model (OSWM)}, a transformer-based synthetic environment trained entirely on synthetic data sampled from a prior distribution of untrained, randomly initialized neural networks. These neural networks serve as the prior, with each one mimicking a specific RL environment dimension. In contrast to SEs, OSWM employs in-context learning and a supervised objective without meta-learning to predict the next states and rewards at random cut-off points that match the synthetic trajectories. Unlike SEs, which are tailored to a single target environment, OSWM provides a general proxy capable of representing multiple environments within a single model. After training the OSWM, it acts as a learned simulator for multiple environments, enabling RL agents to train policies purely on the synthetic dynamics it generates. To adapt OSWM to a new target environment, we sample a context of 1,000 transitions from the environment. This context is sufficient for OSWM to infer the dynamics of the target environment and generate synthetic trajectories for training RL agents. Without requiring further interactions with the target environment, OSWM enables agents to achieve competitive performance in simple control environments such as GridWorld, CartPole, a custom control environment, and mediocre performance on Reacher. We also investigate the effect of context sampling and the role of the prior with respect to OSWM's performance. Our empirical results indicate that with better prior design, OSWM may capture more complex dynamics and scale to more challenging environments.

\textbf{How does this work address the overarching scientific question?}
Our work demonstrates how synthetic data and in-context learning can create a general simulator for multiple RL environments and hereby bridges the core themes of meta-learning and synthetic data in this dissertation. Contrary to meta-learned Synthetic Environments, OSWM avoids the computational cost of bi-level optimization by using synthetic priors. However, this comes at the expense of emergent properties such as efficient agent training and hyperparameter robustness. Despite these trade-offs, OSWM underpins the potential of synthetic data approaches for scenarios where real data is expensive, such as in RL and robotics, and contributes to the recent trend of foundation RL models capable of addressing diverse tasks. While RL served as a cost-effective testbed in our work, the methods and insights have applicability beyond RL, offering a foundation for adoption to other domains. 

\section{List of Publications}
In this section, we provide an overview of all core publications included in this dissertation, along with additional publications, patents, and achievements by the author. Detailed descriptions of the author contributions are provided in the appendix, with references to their specific sections indicated at the beginning of each chapter.

\paragraph{Core Publications} This dissertation integrates the following eight core research papers (sorted in chronological order). All research papers follow the overarching scientific question of \emph{``How can meta-learning and synthetic data advance automated pretraining and finetuning?"}. We also denote code references and presentation types (if awarded).
\begin{refsection}
\begin{itemize}
    \item \fullciteexclude{ferreira-iclr22a-highlighted}\\
    Code: \url{https://github.com/automl/learning_environments}

    \item \fullciteexclude{wagner-icml22a-highlighted}\\
    Code: \url{https://github.com/automl/importance_hp_da_ssl}

    \item \fullciteexclude{ferreira-icml22a-highlighted}\\
    Code: \url{https://github.com/automl/zero-shot-automl-with-pretrained-models}
    
    Note: \textit{Awarded with a spotlight presentation at ICML 2022}.
    \item \fullciteexclude{pineda-iclr24a-highlighted}\\
    Code: \url{https://github.com/machinelearningnuremberg/quicktune} Note: \textit{Awarded with an oral presentation at ICLR 2024}.

    \item \fullciteexclude{ferreira-neuripsws24a-highlighted}\\
    Code: \url{https://github.com/automl/oswm}\\

    \item \fullciteexclude{strangmann-neuripsws24a-highlighted}\\
    Code: see Quick-Tune-Tool code.

    \item \fullciteexclude{rapant-automlws24a-highlighted}\\
    Code: \url{https://github.com/automl/quicktunetool}

    \item \fullciteexclude{ferreira-iclr25a-highlighted}\\
    Code: \url{https://github.com/automl/pretraining-hard-views/}
    
\end{itemize}

\paragraph{Further Publications} 
The following lists additional publications that the author conducted during their doctoral program. While these publications are connected to the ideas presented in this dissertation, they are considered out-of-scope within this dissertation. Additionally, we also list the author's patents that resulted from their works during the program. 

\begin{itemize}
    \item \fullciteexclude{zu-tpami21a-highlighted}
    \item \fullciteexclude{baz-pmlr22a-highlighted}
    \item \fullciteexclude{rajan-jair23a-highlighted}
\end{itemize}

\paragraph{Patents}
\begin{itemize}
    \item \fullciteexclude{ferreira22b-highlighted}
    \item \fullciteexclude{ferreira22a-highlighted}
\end{itemize}

\end{refsection}

\paragraph{Further Achievements}
We competed with an early version of ZAP in the 2019 ChaLearn AutoDL Challenge, focusing exclusively on the vision track. Our submission achieved performance on par with the competition winner on this track, with an average rank of 1.75, as detailed in the corresponding study \citep{baz-pmlr22a}. Furthermore, ZAP \citep{ferreira-icml22a} received a spotlight presentation at ICML 2022\footnote{\url{https://icml.cc/virtual/2022/spotlight/18008}}, granted to 20\% of submitted papers, while Quick-Tune \citep{pineda-iclr24a} was recognized with an oral presentation at ICLR 2024\footnote{\url{https://iclr.cc/virtual/2024/oral/19719}}, a distinction awarded to only 1.2\% of submissions.

\chapter{Meta-Learning}

\section{What is Meta-Learning?}
Generally, \emph{meta-learning}, or \emph{learning to learn} summarizes any type of learning mechanism that leverages prior experience from tasks \citep{schmidhuber-tum87a, bengio-97a, hochreiter-icann01a, vilalta-air02a, thrun-springer12a}. A description of meta-learning can be given by starting from traditional machine learning. In traditional machine learning, we typically train the parameters $\theta$ of a \emph{learner} \(f\) to maximize a learning cost \(\mathcal{L}\) on a specific task $t \in \mathcal{T}$ (or dataset). In contrast, meta-learning can be described as aiming to train a (meta-)learner to generalize across a distribution of tasks \(p(T)\). We call this process meta-learning because it is concerned with optimizing the parameters \(\theta\) of the meta-learner such that it can leverage prior knowledge to tackle new tasks effectively or efficiently. Formally, this can be expressed as

\[
\theta^* = \argmin_{\theta} \mathbb{E}_{t \sim p(\mathcal{T})} \mathcal{L}(f_\theta, t),
\]
where \(p(\mathcal{T})\) represents the distribution of tasks, and \(\mathcal{L}\) measures the task-specific cost. Typically, the closer the training tasks sampled from \(p(\mathcal{T})\) are to a new task \(t_{\text{new}}\), the better the learner \(f_\theta\) can transfer and utilize prior knowledge to tackle \(T_{\text{new}}\). Here, we consider the categorization of meta-learning techniques proposed in \citet[Chap.~2]{vanschoren-automlbook19a}, which clusters meta-learning approaches by the meta-data $p(\mathcal{T})$ they leverage: learning from model evaluations, task properties, or prior models.



\section{Types of Meta-Learning}
\subsection{Learning from Model Evaluations}
Meta-learning from model evaluations entails learning from meta-data that consists of recordings from model evaluations. Consider that these recordings are defined by their hyperparameter configurations $h_i \in \mathcal{H}$ (e.g., hyperparameter or pipeline settings). Moreover, assume that we have a performance matrix $\mathbb{P}$ a matrix of scalar evaluations $P_{i,j} = P(h_i, t_j)$ of configuration $h_i$ on task $t_j$ measured by a defined evaluation metric and usually recorded before meta-training time. The goal in this sub-type of meta-learning is to train a meta-learner using $\mathbb{P}$ that predicts configurations $\mathcal{H}_{new}^*$ for a new task $t_{new}$. There exist two prominent approaches to this: one is to assume no access to any evaluations on $t_{new}$, which results in training a meta-learner $f: \mathcal{H} \times \mathcal{T} \to {h_k^*}$ that returns a set of best configurations on the meta-training data, independent of $t_{new}$ and to choose the top configurations from this set. On the other hand, if evaluations on $t_{new}$ are permissible, one can transfer configurations using a similarity measure to determine how similar evaluations on $t_{new}$ are to evaluations on the prior tasks $t_j$. Suppose evaluations on $t_{new}$ are similar to the ones on $t_j$. In that case, this knowledge can be exploited to yield the most similar configuration directly. Another family of approaches is to train task-wise surrogate models $s_j(h_i) = P_{i,j}$ used jointly with acquisition function in Bayesian frameworks to suggest a new $h_i$.

\subsection{Learning from Task Properties}
Meta-learning can also be achieved by learning about the characteristics of a task. These characteristics are descriptive features which we refer to as \emph{meta-features}. Each task $t_j$ is represented by a vector of $K$ such descriptive features, i.e. $\phi_j \in \Phi \subseteq \mathbb{R}^K$. 
Meta-features provide a straightforward way to quantify task similarity, for instance, by measuring the cosine similarity between the vectors $\phi_i$ and $\phi_j$. This facilitates meta-learning by transferring information from the most similar tasks to a new task $t_{new}$. A common categorization is to distinguish between learned and not learned meta-features. 

Not learned meta-features are typically numerical features extracted directly or indirectly from the dataset. For instance, one can use the number of classes or image resolution (directly) or apply functions like min, max, mean, quartiles, etc., to compute the minimum value of a feature column in tabular data (indirectly). Further processes like normalization, clustering, landmarking, or dimensionality reduction techniques like PCA can be applied. For a more complete introduction and rationale on selecting meta-features, we refer the interested reader to \cite[chap.~2.3]{vanschoren-automlbook19a}. The selection of meta-features is well correlated with the performance of a meta-learner. However, choosing a well-performing set of meta-features is challenging and dependent on the application, as reported in studies such as \citet{bilalli-ijamcs17a}. 

Because the manual selection of meta-features can be ineffective, studies have investigated whether meta-features can also be learned. This is particularly interesting in the computer vision domain. In contrast to the tabular data domain that consists of low-dimensional data like columns and rows, numerical properties of image or video datasets are often limited in quantity and expressiveness (e.g. pixel values). One approach is to use the performance meta-data or other existing meta-features to learn landmark-like meta-features, i.e. to learn functions $f: \Phi \to \Phi'$, or $f:\mathcal{P}\times\mathcal{H} \to \Phi$ \citep[Chap.~2.3]{vanschoren-automlbook19a}.  More recent approaches take the raw dataset (task) as input and use neural networks to generate meta-features, i.e. $f: \mathcal{T} \to \Phi$ \citep{achille-iccv19a, jomaa-dmkd21a} or learn metric spaces for measuring task similarity \citep{snell-nips17a}. 

Once an appropriate set of meta-features has been identified or learned, the relationship between meta-features and performance meta-data can be learned to predict suitable configurations given the meta-features of $t_{new}$, or more formally: $s: \mathcal{P} \times \Phi \to \mathcal{H}$. The principle of jointly exploring the relationship between meta-features, performance meta-data, and hyperparameter configurations forms this dissertation's central methodological foundation, which we describe in more detail in Section \ref{sec:as_cash} below. A key aspect of this foundation is to leverage transfer learning in neural networks, which we will introduce next.

\subsection{Learning from Prior Models}
\label{sec:learningpriormodels}
Another type of meta-learning revolves around leveraging previously trained machine learning models. Formally, let \( M \) denote a pretrained model whose parameters \(\theta\) have been optimized on one or more tasks \( t_j \sim p(\mathcal{T}) \), drawn from a distribution over tasks. The general idea is to transfer \( M \) to a new task \( t_{new} \sim p(\mathcal{T}) \), using the knowledge encoded in \(\theta\) from the source tasks \(\{ t_j \}\) as a starting point. This process is widely known as \emph{transfer learning} \citep{caruana-multitasklearning, pan-tkde10a, thrun-springer12a, bengio-tpami13a}. While this has been explored for different kinds of machine learning models, neural networks are particularly well-suited due to their model parameters and modular structure. For instance, a standard approach is to replace the final task-dependent layer of a neural network and reuse the remaining pretrained layers as feature extractors, thereby providing a useful initialization that can be finetuned $t_{new}$\citep{pratt-nips92a, thrun-ijcai95a, thrun-nips95a, bengio-icml11a}. 

A body of literature that takes advantage of transfer learning is \emph{few-shot learning}, where the idea is to adapt \( M \) to a new task using only a few training examples. Through meta-learning, bootstrapping the initialization such that a common feature representation serves as an inductive bias enables rapid adaptation of the latent representation to \( t_{new} \) from only a few examples \citep{ravi-iclr17a, finn-icml17a}. Similar ideas have also been realized through memory-augmented neural networks that learn to retrieve and leverage relevant ``memories" from previously seen tasks \citep{santoro-icml16a}. Moreover, another branch of approaches leveraging transfer learning focuses on meta-learning algorithms within neural networks, for instance, by learning optimizers that adapt their own weights or hyperparameters to new tasks \citep{schmidhuber-92a, andrychowicz-neurips16a, li-iclr17b}. For other approaches that use meta-learning to leverage prior trained models, we refer the interested reader to \citep[Chap.~2.4]{vanschoren-automlbook19a}.

Next, we introduce a framework that allows meta-learning of the relationship between meta-features, performance meta-data, and hyperparameter configurations and adapt it to automate the finetuning of deep learning pipelines by leveraging transfer learning.

\section{Algorithm Selection and CASH} \label{sec:as_cash}
In the context of Machine Learning, selecting the most appropriate algorithm (e.g., classifiers or regression models) from a discrete set of algorithms without assuming to be able to make observations on the target \emph{task} is known as the \emph{Algorithm Selection} problem~\citep{rice-aic76a,smithmiles-acmcs08a,kotthoff-aicom12a,bischl-aij16a}. Below, we formalize it and extend it to the \emph{Combined Algorithm Selection and Hyperparameter Optimization (CASH)} problem~\citep{thornton-kdd13a, hutter-book19a}.

Given a set of algorithms \(\mathcal{A}\) and tasks \( t \in \mathcal{T} \) (also referred to as \emph{problem instance} and often represented by meta-features), along with a cost metric \( c: \mathcal{A} \times \mathcal{T} \to \mathbb{R} \), the goal is to learn a mapping $s: \mathcal{T} \to \mathcal{A}$ which is often referred to as a \emph{selector} or surrogate model and that minimizes the total cost:

\begin{equation}
s^* = \argmin_{s: \mathcal{T} \to \mathcal{A}} \sum_{t \in \mathcal{T}} c\bigl(s(t), t\bigr).
\end{equation}

Here, the cost metric \( c\bigl(s(t), t\bigr) \) evaluates the performance of algorithm \( s(t) \) on task \( t \). Because no individual machine learning model typically achieves the best performance across all tasks, and many machine learning problems require hyperparameter optimization to achieve good performance, it is natural to extend hyperparameter optimization to treat the choice of algorithms as a hyperparameter. The CASH problem integrates hyperparameter optimization with algorithm selection. This extension considers not only the selection of the optimal algorithm for each task but also the tuning of its hyperparameters to minimize the total cost.

To formalize CASH, let \(\mathcal{H}_i\) denote the hyperparameter space associated with each algorithm \( A_i \in \mathcal{A} \). The combined algorithm and hyperparameter space is given by:
\[
S = \bigcup_{i=1}^{n} \{A_i\} \times \mathcal{H}_i,
\]
where each element \((A_i, h_i) \in S\) represents a specific algorithm \( A_i \) and a configuration of its hyperparameters \( h_i \in \mathcal{H}_i \). The performance of \( s(t) \) on task \( t \) is now extended to include hyperparameters, expressed as:
\[
c\bigl((s(t), h_i), t\bigr).
\]

The objective of CASH is to jointly optimize the choice of algorithm and its hyperparameters to minimize the total cost across all tasks:

\begin{equation}
(s^*, h^*) = \argmin_{s: \mathcal{T} \to \mathcal{A}, \ h_i \in \mathcal{H}_i} \sum_{t \in \mathcal{T}} c\bigl((s(t), h_i), t\bigr).
\label{eq:cash}
\end{equation}

For more details on how to solve Equation \ref{eq:cash} with mostly applications outside of Deep Learning (DL), we refer the interested reader to \citet{thornton-kdd13a}, \citet{hutter-book19a}, and \citet{feurer-jmlr22a}. 

In this dissertation, we adapt the CASH framework to DL pipelines, where the combined algorithm and hyperparameter tuple correspond to pretrained models \( M_i \) and their finetuning hyperparameters \( h_i \), i.e., \( x = (M_i, h_i) \) and the cost metric is the loss (e.g., cross-entropy loss in supervised learning). In this setting, a dataset represented by its meta-features \(\phi \in \Phi\) is typically split into training, validation, and test splits, i.e., \( D^{(tr)}, D^{(val)}, D^{(test)} \). The cost is then defined as:
\[
c(x, \phi) = \mathcal{L}\bigl(\mathrm{Tune}(x, D^{(tr)}, D^{(val)}), D^{(test)}\bigr),
\]
where \(\mathcal{L}\) measures the loss on the test split after finetuning the pretrained model on the training split.

Analogously to the CASH setting, we aim to optimize the objective given by Equation \ref{eq:cash} by training the selector or surrogate model on meta-data. In the case of DL pipelines, this meta-data consists of prior recorded performance data about pretrained models \( M_i \) when finetuned on similar tasks (identified by meta-features \(\phi_i\)) given a finetuning hyperparameter configuration \( h_i \). Once trained, the surrogate can make informed decisions about which DL pipeline to select for new tasks \(\phi_{\text{new}}\), depending on how well the data on which the surrogate was trained resembles future tasks. In this dissertation, we explore learning both probabilistic surrogates that model uncertainty and non-probabilistic surrogates that predict point estimates to select DL pipelines.

\chapter{Data Augmentation and Synthetic Data}
Data augmentation and synthetic data generation resemble essential methods to systematically expand, modify, or produce entirely new datasets \( D_t \) for a given task \( t \). By increasing data diversity, these methods often allow for improving model robustness and mitigate overfitting \citep{image_augmentation_survey}. In this chapter, we begin by introducing data augmentation and synthetic data generation in a general and conceptual manner, highlighting their core principles. We then narrow our focus to a subset of approaches that are most relevant to the scope of this thesis.

\section{Data Augmentation}
\begin{wrapfigure}[19]{r}{0.4\textwidth}
    \vspace{-0.3cm}
    \centering
    \includegraphics[width=0.4\textwidth]{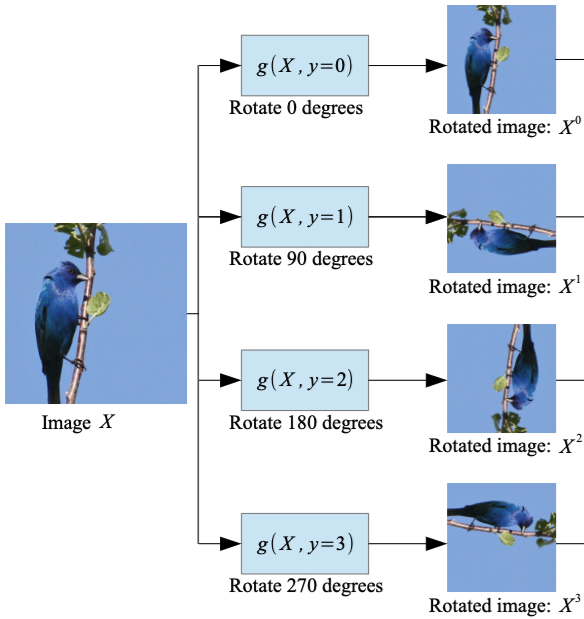}
    \caption{Predicting image rotations instead of classes in Self-Supervised Learning (visualization from \citet{gidaris-iclr18a}).}
    \label{fig:rotation}
\end{wrapfigure}

Data augmentation applies transformations to existing data. Assume \(\mathcal{F}\) is a set of sample level transformations. Each transformation \(f \in \mathcal{F}\) maps \( D_t \) to an augmented dataset \(\tilde{D}_t\). On a sample level, data augmentation transforms each sample \( x \in D_t \) independently. In computer vision, these transformations are typically composed of geometric (cropping, flipping, rotating, cut-out, etc.) or appearance (color distortion, blurring, etc.) perturbations \citep{image_augmentation_survey}. Formally, this can be described as \(\tilde{D}_t = \{f(x) \mid x \in D_t, f \sim p(f \mid \mathcal{F})\}\). Data augmentation can also occur at the dataset level, where a transformation modifies the characteristics of the entire dataset, for instance, changing the number of samples or the resolution of images.

Manual augmentation strategies often rely on domain-specific prior knowledge. To automate the augmentation process, automated data augmentation techniques have been proposed and nowadays resemble important components of modern learning pipelines. In computer vision, for instance, learning-based approaches such as AutoAugment (AA) \citep{cubuk-cvpr19a}, Population-based Augmentation (PBA) \citep{ho-icml19a}, or Fast AutoAugment (FAA) \citep{sungbin-neurips19a} aim to learn dataset specific augmentation policies. AA treats augmentation as a sequential decision problem and uses reinforcement learning to solve it, however, at a high computational cost. PBA reduces this cost by learning multiple worker schedules via evolution-based training. Further reducing the computational cost, FAA employs a strategy that identifies good sub-policies on dataset splits. On the other hand, works exist that learn augmentation networks to generate pixel-level augmentations with an adversarial objective \citep{antoniou2017data}. However, there also exist pure sampling-based approaches like RandAugment \citep{cubuk-cvpr20a} and TrivialAugment \citep{muller-iccv21a} which are cheaper than the learning-based approaches since they simplify this process further by eliminating the need for any additional learning augmentation policies or generators. Instead, they directly randomly sample the number and magnitude of transformations and show comparable performance to the learning-based approaches.\newline

\begin{figure}[t]
    \centering
    \begin{minipage}[b]{0.5\textwidth}
        \centering
        \includegraphics[width=\textwidth]{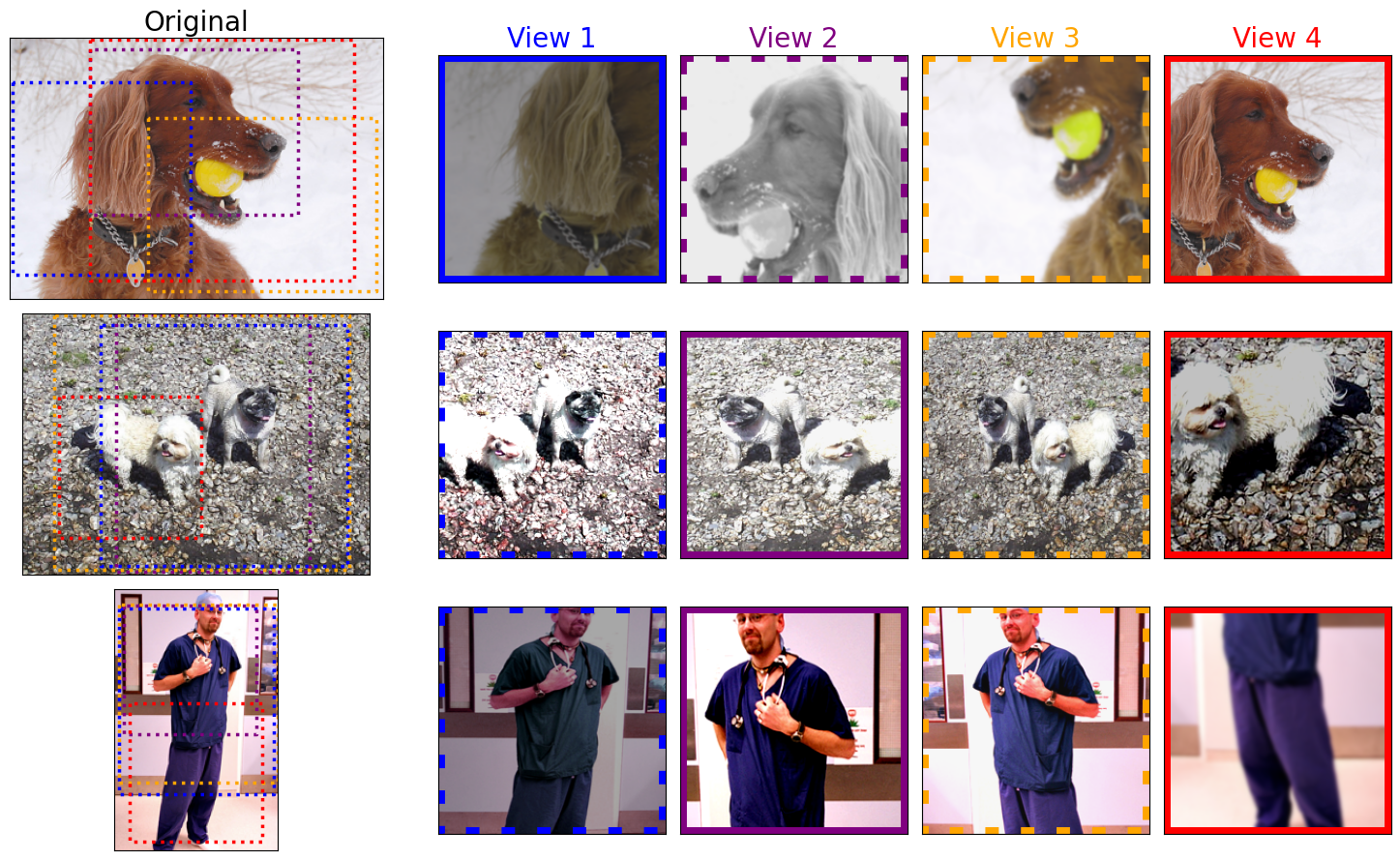}
        \label{fig:example_views}
    \end{minipage}
    \hfill
    \begin{minipage}[b]{0.4\textwidth}
        \centering
        \includegraphics[width=\textwidth]{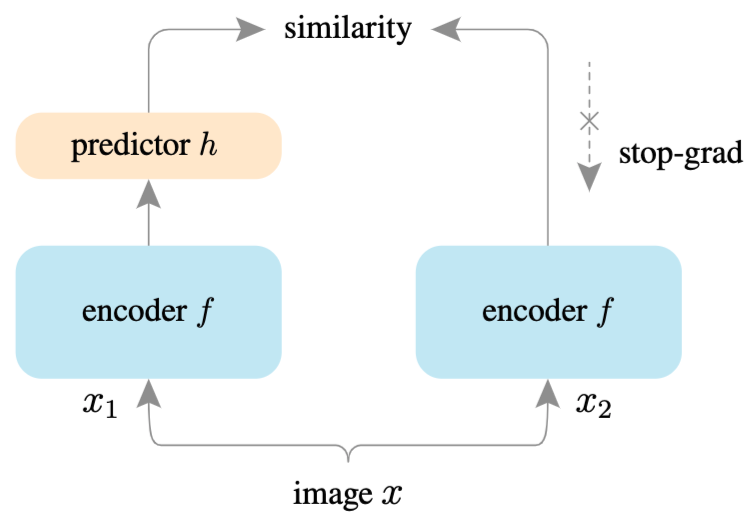}
        \label{fig:simsiam}
    \end{minipage}

    \caption{\textbf{Left:} Example views after applying geometric and appearance 
    transformations (visualization from \citet{ferreira-iclr25a}). \textbf{Right:} SimSiam's architecture asymmetry to facilitate contrastive learning 
    without negative image pairs (visualization from \citet{chen-cvpr21b}).}
    \label{fig:views_simsiam}
\end{figure}

Data augmentation is particularly essential in self-supervised learning (SSL), where models learn representations without explicitly requiring a labeled dataset by directly extracting a supervision signal from the data. Early SSL approaches formulate and optimize the task of predicting transformations applied to images. For instance, predicting the spatial (pixel) context of image patches \citep{doersch-iccv15a}, reconstructing compressed representations back into pixel-space \citep{bourlard1988auto, hinton1993autoencoders}, solving Jigsaw puzzles, where images are divided into patches and shuffled \citep{noorozi-eccv16a}, or classifying randomly sampled rotations applied to images \citep{gidaris-iclr18a} (see Fig. \ref{fig:rotation}). By forcing the model to predict or recover the applied transformations, these methods encourage the model to learn semantically meaningful features.\newline

This methodology is also the foundation for the more recent and popular \emph{contrastive} SSL framework \citep{chen-icml20b} which focuses on learning representations by learning to distinguish between different augmented inputs, referred to as \emph{views} (see Fig. \ref{fig:views_simsiam} left). This framework learns latent representations in which similar image views are located closely, and dissimilar ones distantly. However, for effective learning in this framework, a large pool of negative samples is required to prevent the model from collapsing to trivial solutions \citep{chen-icml20b}. More recent methods, such as BYOL \citep{grill-neurips20a}, SimSiam \citep{chen-cvpr21b}, DINO \citep{caron-iccv21a}, and iBOT \citep{zhou-icml22a} eliminate explicit negatives by introducing additional architectural components, such as architectural asymmetry, momentum encoders or averaging schemes, that prevent collapse while relying solely on positive pairs.\\

More formally, we describe the contrastive learning objective in the example of SimSiam~\citep{chen-cvpr21b}, 
which is also visualized in Fig. \ref{fig:views_simsiam} (right). 
Let \(\mathcal{D}\) be a set of images, and \(\mathcal{F}\) a set of sample level transformations. 
Consider a minibatch of \(M\) images \(\mathbf{x} = \{x_i\}_{i=1}^M\) sampled uniformly from \(\mathcal{D}\). 
SimSiam draws two random transformations \(f_1, f_2 \sim p(f \mid \mathcal{F})\) and applies them to each image in \(\mathbf{x}\), 
resulting in two augmented sets of views \(\mathbf{x}^1\) and \(\mathbf{x}^2\). An encoder \(f_{\theta}\) and a predictor \(h_{\theta}\) then produce embeddings. For the first view, we define 
\[
\mathbf{e}^1 = f_{\theta}(\mathbf{x}^1), 
\quad 
\mathbf{z}^1 = h_{\theta}(\mathbf{e}^1),
\]
with analogous definitions \(\mathbf{e}^2\) and \(\mathbf{z}^2\) for the second view. 
SimSiam then minimizes
\begin{equation}
\label{eq:simsiam}
\mathcal{L}(\theta) 
= 
\frac{1}{2} 
\Bigl(
D(\mathbf{z}^1, \mathbf{e}^2) 
+
D(\mathbf{z}^2, \mathbf{e}^1)
\Bigr),
\end{equation}
where \(D(\cdot,\cdot)\) denotes negative cosine similarity. Unlike standard contrastive frameworks, SimSiam does not rely on negative examples. Instead, it avoids trivial solutions (all embeddings collapsing to a constant) by introducing \emph{asymmetry} in its architecture and objective by introducing a small network (predictor) to only one branch and restricting gradient flow on the other. By learning to align these slightly different outputs, SimSiam maintains non-trivial embeddings. A more recent body of literature learns such embeddings through adversarial objectives to automatically select more challenging views \citep{ kocyigit-wacv23, ferreira-iclr25a} that aid SSL training or optimize augmentation networks to directly output augmented views given images \citep{tian-neurips20a, shi-icml22a, tamkin-iclr21a}. Collectively, these approaches share the core principle of aligning representations across augmentations of the same input, which enables improved downstream task performance without requiring labeled data.

\section{Synthetic Data Generation}
In contrast to data augmentation, synthetic data generation is concerned with creating entirely new data samples or datasets.
One way to employ synthetic data generation is to learn a data-generating model \( G_\psi \) trained on one or more existing tasks to produce \( D_t = \{ x \mid x \sim G_\psi \} \). Many different approaches have been proposed. A popular method is Generative Adversarial Networks (GANs) \citep{goodfellow-nips14a}. GANs consist of two neural networks, a generator and a discriminator, that are optimized in a joint adversarial learning process. While the generator creates synthetic data samples that mimic the real data, the discrimination distinguishes between real and synthetic samples. Through optimization, the generator's goal is to effectively fool the discriminator, and the discriminator enhances its ability to detect synthetic samples. Post optimization, the resulting generator can be used as \( G_\psi \) to produce new samples independent of the real data-generating process. In addition to GANs, many other generative model techniques exist. For instance, Variational Autoencoders use probabilistic modeling with an encoder that projects data into a latent space and a decoder that reconstructs it, where the generation function \( G_\psi \) samples from the latent distribution and decodes it back to the original space. Flow-based models and Diffusion models are complementary techniques to learn \( G_\psi \). They too learn latent spaces from which one can sample synthetic data. Flow-based models do this by learning an invertible mapping from data to latent space through a sequence of transformations. On the other hand, diffusion models learn a reversion function that iteratively denoises random additive noise to arrive at a (synthetic) data sample.\\

Typically, works using these techniques are incentivized to mimic the real data distribution. However, there exist approaches that use learning objectives that facilitate more efficient learning when training on the produced synthetic data. Particular examples are Generative Teaching Networks \citep{such-icml20a} that meta-learn \( G_\psi \) to generate synthetic data specifically optimized to accelerate the learning process of a target model: the inner loop trains the target model on the synthetic data produced by \( G_\psi \), and an outer loop evaluates the target model on a real target task. As illustrated in \Cref{fig:cash_synthetic_data} (right), the meta gradients are then backpropagated through the entire learning process to update \( G_\psi \). A related approach is dataset distillation \citep{wang-arxiv18a}, which backpropagates the validation error gradients directly into the input to generate a synthetic dataset to achieve comparable or better performance than training on the full dataset. In a similar fashion, \citet{maclaurin-icml15a} optimizes directly for the data by treating it as hyperparameters. Another technique for improving training efficiency is core-set selection \citep{tsang-icml05a, sener-iclr18a}. Here, a subset of the original dataset is identified as a proxy dataset which encompasses the most relevant information needed for effective learning. Together, these approaches demonstrate various strategies for generating synthetic datasets, targeting objectives such as efficient learning, reduced dataset size, or the ability to produce large amounts of synthetic data tailored for specific tasks.\\

A different methodology is to sample data from a user-specified or prior known distribution over data points and labels \( q(\mathcal{X}, \mathcal{Y}) \), resulting in \( D_t = \{ x,y \mid x,y \sim q(x,y) \} \) and providing a source of synthetic data that does not depend on the original dataset \( D_t \) at all. For example, Prior-Data Fitted Networks (PFNs) \citep{muller-iclr22a, hollmann-iclr23a} utilize this concept by defining a synthetic prior over supervised learning tasks. These priors encode assumptions about the data-generating process (e.g., the relationships between features or causal structures) and enable the generation of synthetic datasets tailored to the task domain. PFNs repeatedly draw a dataset from the synthetic prior distribution, sample data points and their labels from it, mask one of the labels, and optimize the PFN to predict the masked label given the rest of the data points and labels. By training entirely on synthetic data generated from priors and achieving state-of-the-art performance, PFNs showcase the surprising ability to transfer synthetic pretraining to real-world tasks. Backed by the possibility of generating infinite amounts of synthetic data, PFNs and the synthetic data generation methodology in general offer a powerful framework for scaling up models as training is decoupled from the limitations of real-world datasets.



\part{Meta-Learning for Automated Model Selection and Finetuning}
\label{part:meta-learning-cash}


\chapter{Zero-Shot AutoML with Pretrained Models}
\label{chap:zap}

The content of this chapter has been published as:
\highlightfullcite{ferreira-icml22a}

This work represents a core contribution to this dissertation and was published in a peer-reviewed A* conference (CORE2023), with a significant author contribution. The supplementary material and a detailed statement of contributions is provided in Appendix \ref{appendix:sup_zap}.

\includepdf[pages=1-13,offset=0 -1.25cm,addtotoc={
     1,section,1,Introduction,zap_p1,   
     2,section,1,Related Work,zap_p2,
     3,section,1,Zero-Shot AutoML with Pretrained Models,zap_p3,
     5,section,1,ZAP Meta-Dataset Design,zap_p4,
     6,section,1,Experiments,zap_p5,
     9,section,1,Conclusion,zap_p6,
     9,section,1,Limitations,zap_p7
     }]{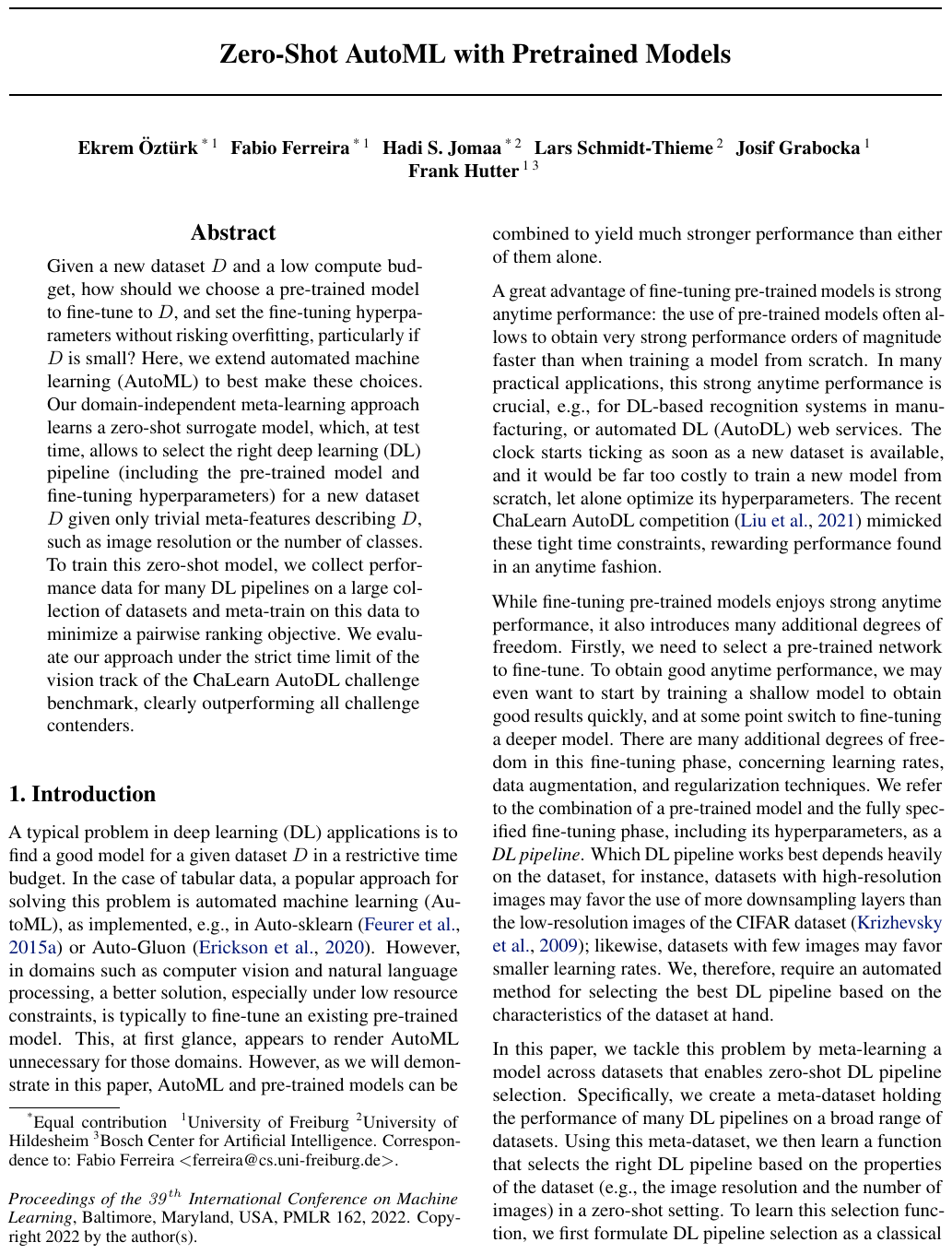}

\chapter{Quick-Tune: Quickly Learning Which Pretrained Model to Finetune and How}
\label{chap:quicktune}

The content of this chapter has been published as:
\highlightfullcite{pineda-iclr24a}

The supplementary material and a detailed statement of contributions is provided in Appendix \ref{appendix:sup_qt}.

\includepdf[pages=1-14,offset=0 -1.25cm,addtotoc={
     1,section,1,Introduction,qt_p1,   
     2,section,1,Related Work,qt_p2,
     3,section,1,Motivation,qt_p3,
     3,section,1,Quick-Tune: Cost-Efficient Finetuning,qt_p4,
     5,section,1,Quick-Tune: Meta-Dataset,qt_p5,
     6,section,1,Experiments and Results,qt_p6,
     9,section,1,Conclusion,qt_p7
     }]{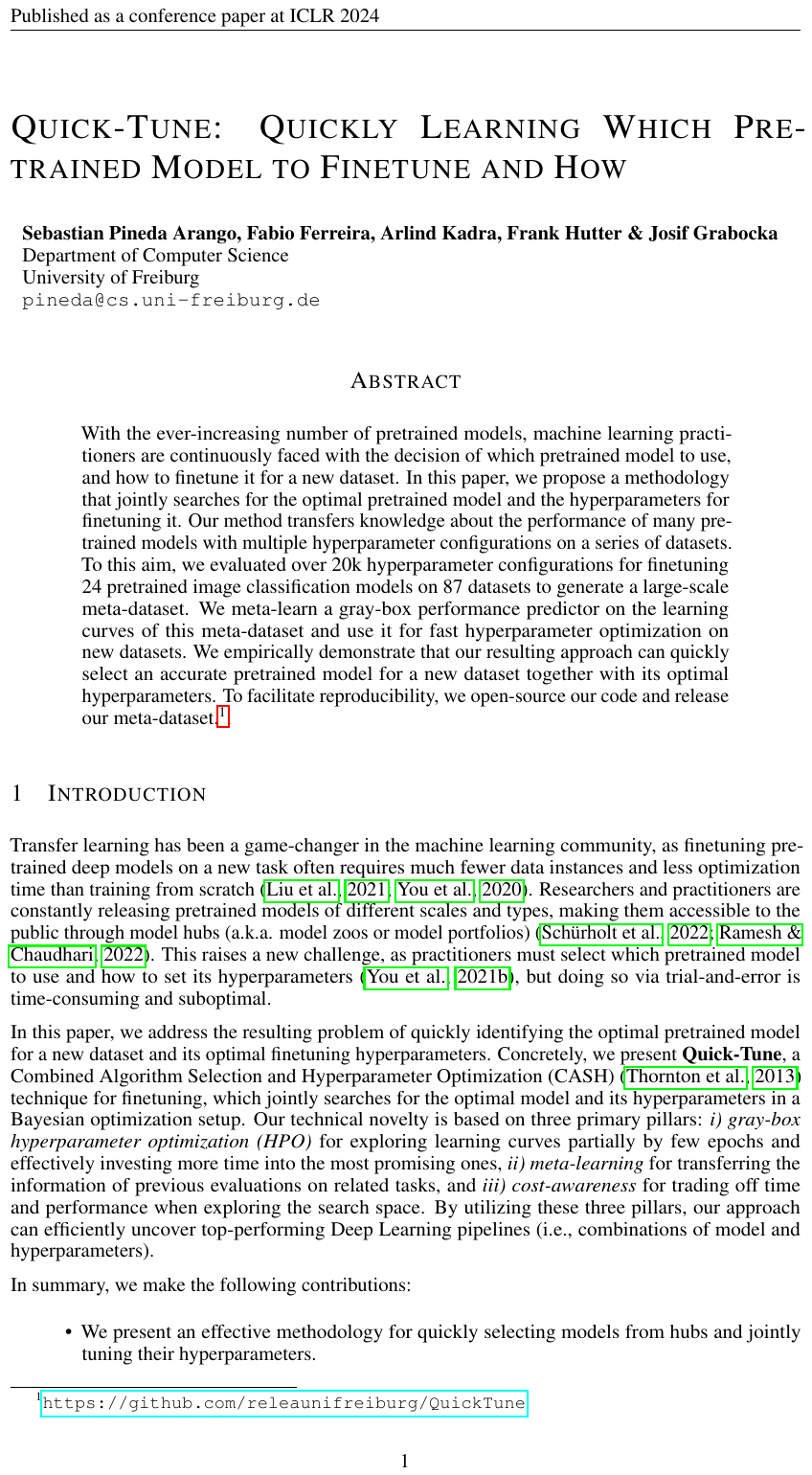}

\chapter{Quick-Tune-Tool: A Practical Tool and its User Guide for Automatically Finetuning Pretrained Models}

\label{chap:qtt}

The content of this chapter has been published as:
\highlightfullcite{rapant-automlws24a}

The supplementary material and a detailed statement of contributions is provided in Appendix \ref{appendix:sup_qtt}.

\includepdf[pages=1-8,offset=0 -1.25cm,addtotoc={
     1,section,1,Introduction,qtt_p1,   
     2,section,1,Background and Related Work,qtt_p2,
     2,section,1,Quick-Tune-Tool: A Practical Tool for Finetuning Pretrained Models,qtt_p3,
     4,section,1,A User Guide for Quick-Tune-Tool,qtt_p4,
     4,section,1,Expriments and Results: Quick-Tune-Tool in Action,qtt_p5,
     5,section,1,Conclusion and Outlook,qtt_p6
     }]{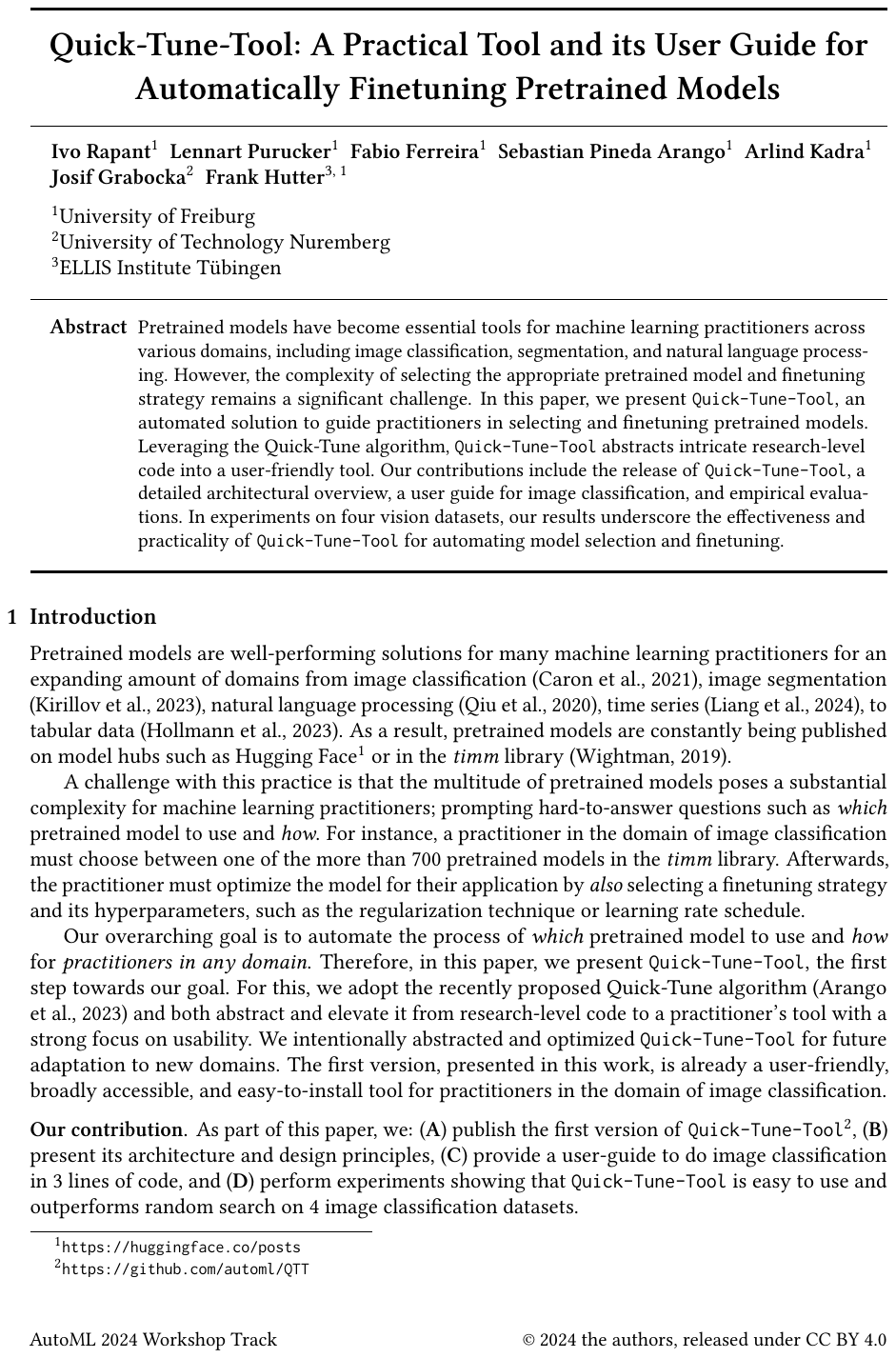}
\chapter{Transfer Learning for Finetuning Large Language Models}\label{chap:qt_llms}

The content of this chapter has been published as:
\highlightfullcite{strangmann-neuripsws24a}

The supplementary material and a detailed statement of contributions is provided in Appendix \ref{appendix:sup_qt_llms}.

\includepdf[pages=1-7,offset=0 -1.25cm,addtotoc={
     1,section,1,Introduction,qt_llms_p1,   
     2,section,1,Related Work,qt_llms_p2,
     2,section,1,Method,qt_llms_p3,
     4,section,1,Results,qt_llms_p4
     }]{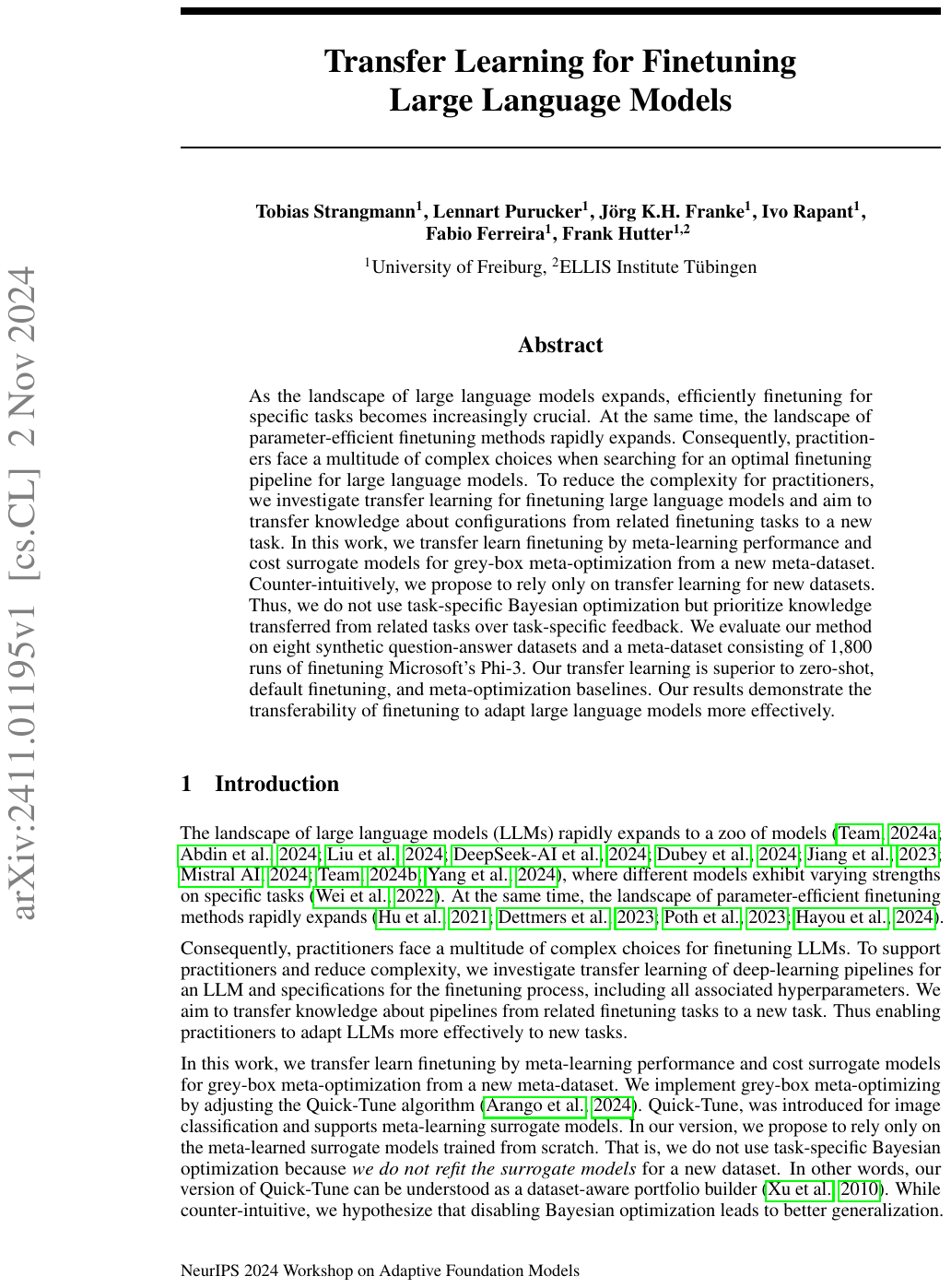}

\part{Meta-Learning Data Augmentation and Synthetic Data for Enhanced Learning}
\label{part:meta-learning-synthetic-data}

\chapter{On the Importance of Hyperparameters and Data Augmentation for Self-Supervised Learning}\label{chap:on_the_importance_ssl}
The content of this chapter has been published as:
\highlightfullcite{wagner-icml22a}

The supplementary material and a detailed statement of contributions is provided in Appendix \ref{appendix:sup_on_importance_ssl}.

\includepdf[pages=1-6,offset=0 -1.25cm,addtotoc={
     1,section,1,Introduction,on_p1,   
     1,section,1,Background and Related Work,on_p2,
     2,section,1,Study on the Importance of Hyperparameters and Data Augmentation,on_p3,
     3,section,1,GroupAugment,on_p4,
     4,section,1,Conclusion and Limitatons,on_p5
     }]{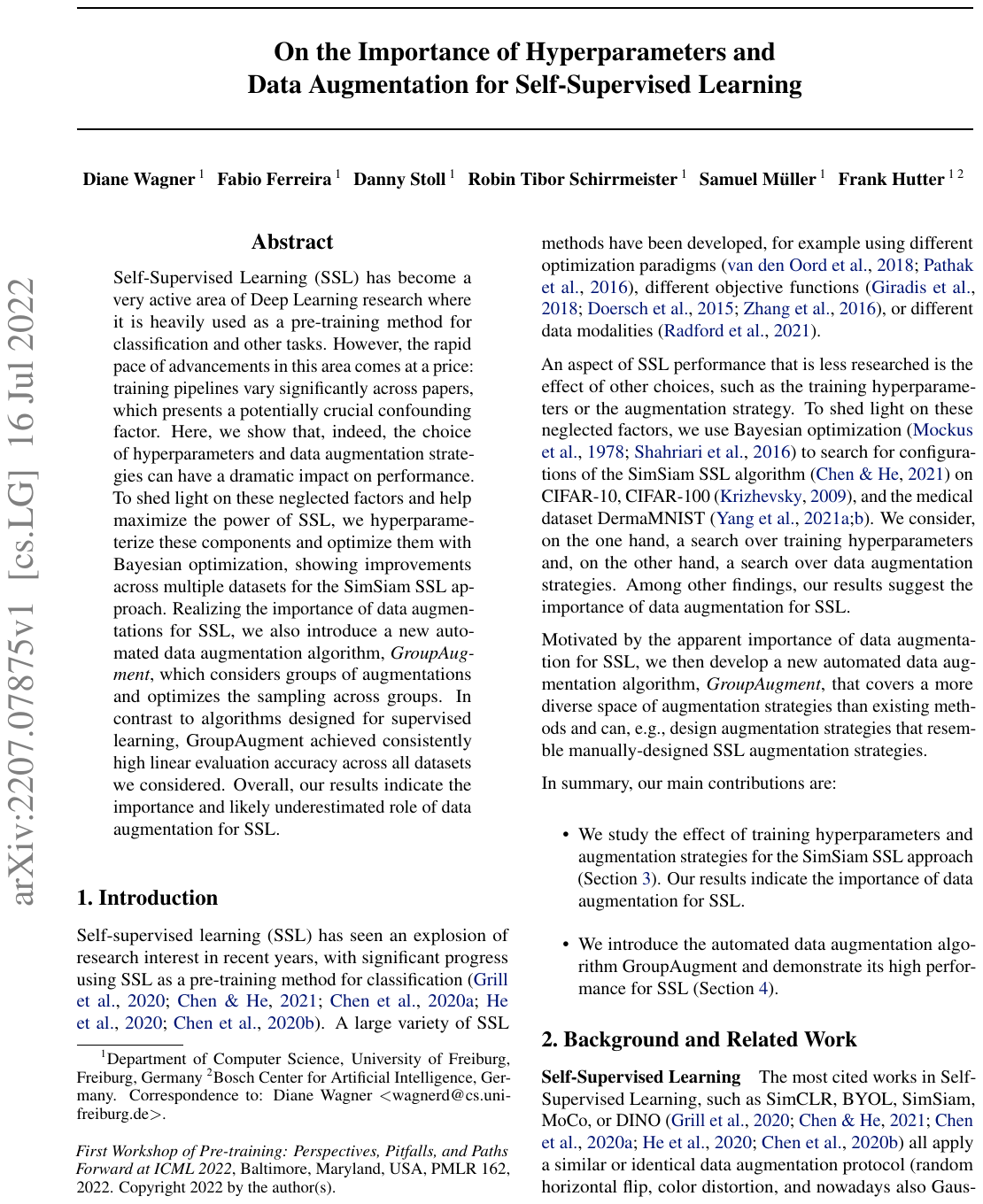}

\chapter{Beyond Random Augmentations: Pretraining with Hard Views}\label{chap:hvp}
The content of this chapter has been published as:
\highlightfullcite{ferreira-iclr25a}

The supplementary material and a detailed statement of contributions is provided in Appendix \ref{appendix:sup_hvp}.

\includepdf[pages=1-12,offset=0 -1.25cm,addtotoc={
     1,section,1,Introduction,hvp_p1,   
     3,section,1,Related Work,hvp_p2,
     3,section,1,Method,hvp_p3,
     5,section,1,Implementation and Evaluation Protocols,hvp_p4,
     5,section,1,Main Results,hvp_p5,
     7,section,1,Empirical Analysis of HVP,hvp_p6,
     9,section,1,Conclusion,hvp_p7
     }]{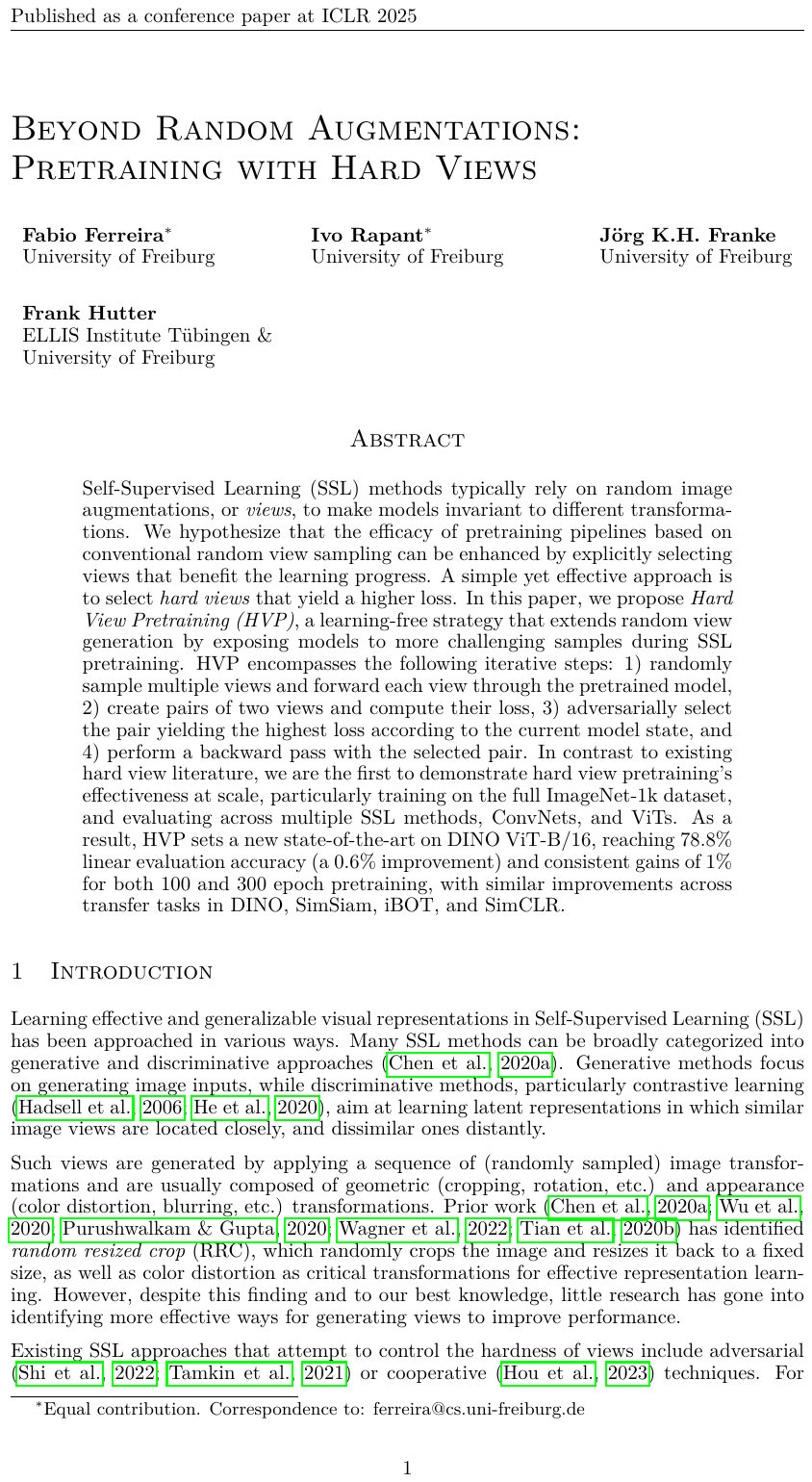}
\chapter{Learning Synthetic Environments and Reward Networks for Reinforcement Learning}\label{chap:learningses}
The content of this chapter has been published as:
\highlightfullcite{ferreira-iclr22a}

The supplementary material and a detailed statement of contributions is provided in Appendix \ref{appendix:sup_se}.

\includepdf[pages=1-12,offset=0 -1.25cm,addtotoc={
     1,section,1,Introduction,ses_p1,   
     2,section,1,Related Work,ses_p2,
     3,section,1,Learning Synthetic Environments,ses_p3,
     4,section,1,Learning Reward Networks,ses_p5,
     5,section,1,Experiments with Synthetic Environments,ses_p6,
     7,section,1,Experiments with Reward Networks,ses_p9,
     9,section,1,Limitations,ses_9,
     9,section,1,Conclusion,ses_10
     }]{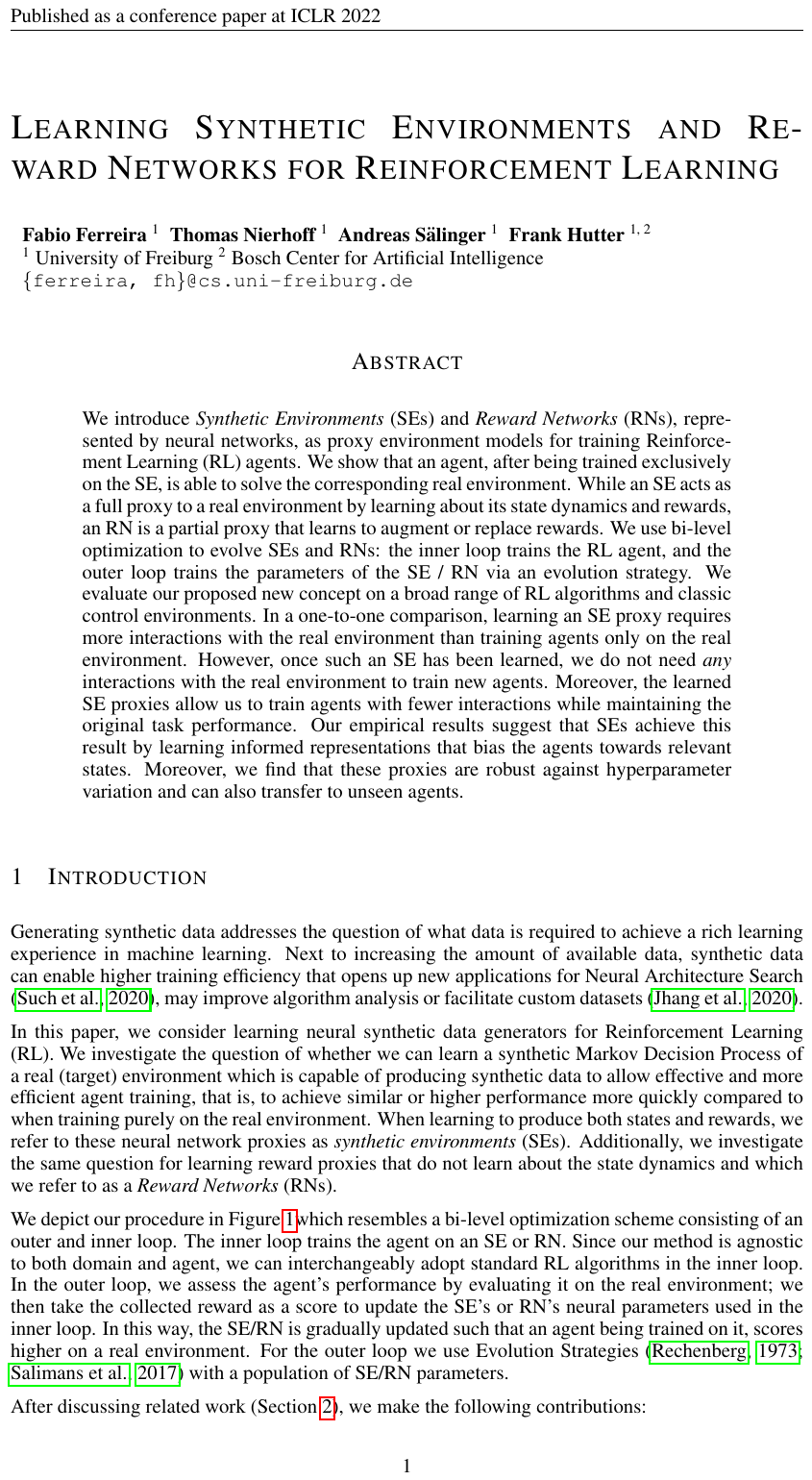}
\chapter{One-shot World Models Using a Transformer Trained on a Synthetic Prior
}\label{chap:oswm}
The content of this chapter has been published as:
\highlightfullcite{ferreira-neuripsws24a}

The supplementary material and a detailed statement of contributions is provided in Appendix \ref{appendix:sup_oswm}.

 \includepdf[pages=1-11,offset=0 -1.25cm,addtotoc={
     1,section,1,Introduction,oswm_p1,   
     2,section,1,Related Work,oswm_p2,
     3,section,1,Method,oswm_p3,
     5,section,1,Experiments,oswm_p4,
     9,section,1,Conclusion,oswm_p6
     }]{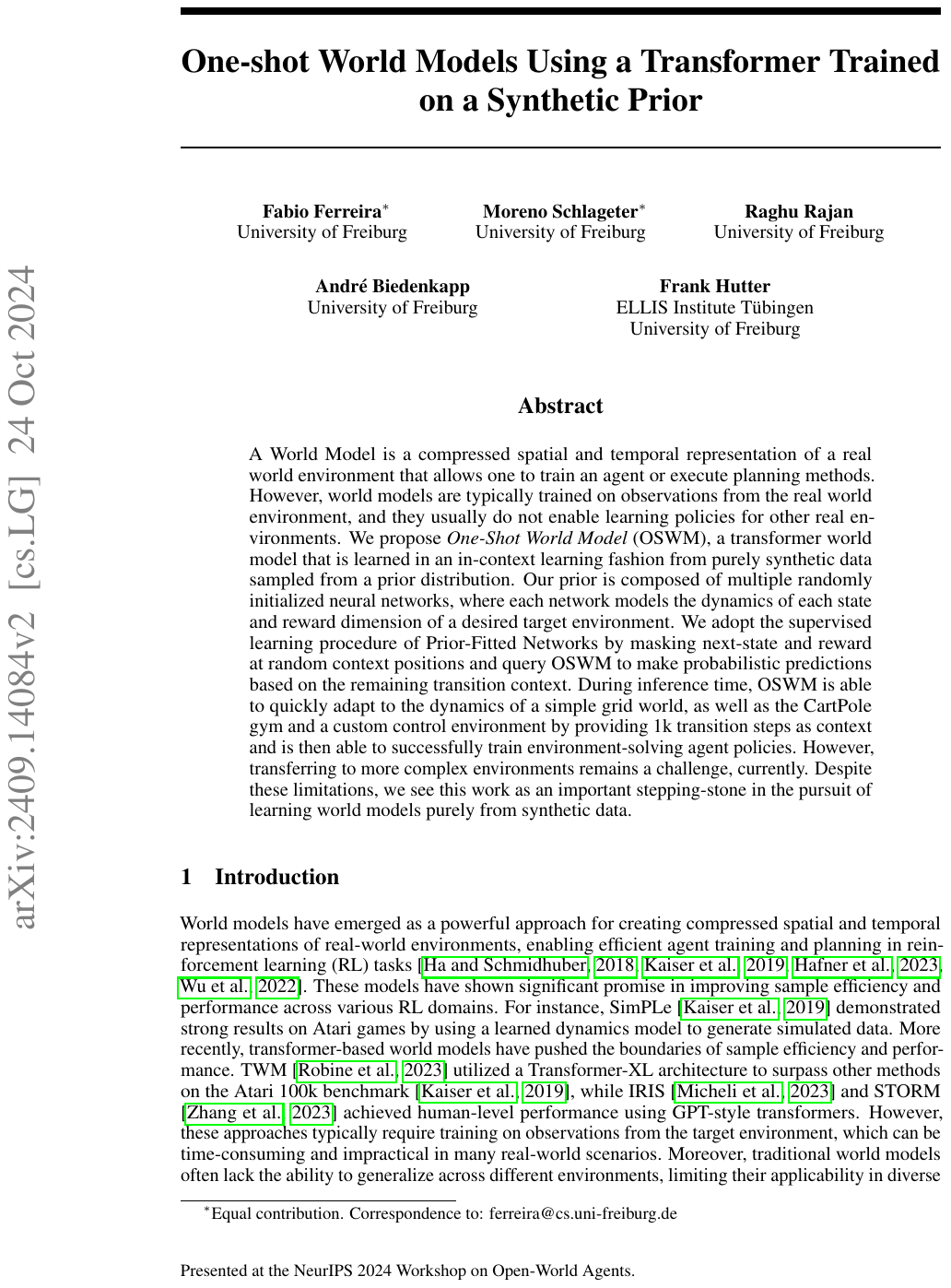}
\part{Conclusion}\label{part:conclusion}

\chapter{Summary and Discussion}\label{chap:summary}
In this chapter, we will now summarize the contributions presented in this dissertation. In \Cref{part:intro}, we outlined the broader context and motivation for the dissertation topic and derived three key challenges this work sought to answer:\\

\begin{tabular}{@{}lp{0.8\linewidth}}
    \textbf{Challenge 1:} & Automating Model Selection and Finetuning Amidst a Growing Landscape of Pretrained Models \\
    \textbf{Challenge 2:} & Enhancing Learning With Advanced Data Augmentation and Synthetic Data Generation \\
    \textbf{Challenge 3:} & Ensuring Reproducibility and Practical Applicability of Automated Learning Methods \\
\end{tabular}

From this, we identified the overarching scientific question \emph{``How can meta-learning and synthetic data advance automated pretraining and finetuning?"}. We addressed the first two challenges in the two core parts of this dissertation:\\

\begin{tabular}{@{}lp{0.8\linewidth}@{}}
    \textbf{\Cref{part:meta-learning-cash}:} & \nameref{part:meta-learning-cash} \\
    \textbf{\Cref{part:meta-learning-synthetic-data}:} & \nameref{part:meta-learning-synthetic-data} \\
\end{tabular}\\

The third challenge was a recurring theme throughout the dissertation since the majority of the papers integrated replicable experiments, open-sourcing of code, and transparent reporting for reproduction. In particular, Quick-Tune-Tool directly addressed this challenge by offering a tool to facilitate reproducible experimentation.

Addressing the first key challenge in \Cref{part:meta-learning-cash}, the associated works focus on leveraging meta-learning to automate the selection and hyperparameter optimization of Deep Learning (DL) pipelines for finetuning given a set of candidate pretrained models. The primary emphasis of these works lies in the automation of finetuning computer vision models, and both data augmentation and synthetic data play a secondary role. The core works to highlight in this part are ZAP (\Cref{chap:zap}) and Quick-Tune (\Cref{chap:quicktune}), which propose to address the key challenge by meta-learning zero-shot and few-shot surrogate models based on prior finetuning data. These works' central novelty and contribution showcase how the classic perspective of algorithm selection can be adopted and elegantly extended to provide effective solutions for the deep learning era, allowing the incorporation of hyperparameters such as data augmentation or finetuning strategies. While ZAP employs dataset level augmentation to enrich its performance matrix, showing how data augmentation can be utilized in the input space to aid meta-learning, Quick-Tune exploits data augmentation in its search space. Importantly, the Quick-Tune results empirically illustrate that a suitable model's customization can compete with or outperform the finetuning of large-scale backbones. This underscores the effectiveness of the broader approach to automating DL pipeline selection and customization as a viable strategy for optimizing deep learning models. The remaining works (\Cref{chap:qtt} and \Cref{chap:qt_llms}) in this part mainly show the utility of developing Quick-Tune-Tool to facilitate wider adoption of Quick-Tune and how it can be transferred to automating the finetuning of large language models (LLMs). Like ZAP's dataset augmentation in the input space, \Cref{chap:qt_llms} shows how generating synthetic Q\&A datasets similarly serves as an input-level augmentation strategy for aiding meta-learned LLM finetuning.\newline

In \Cref{part:meta-learning-synthetic-data}, we address the second key challenge of exploring new ways to leverage data effectively through the help of advanced data augmentation and synthetic data generation. In contrast to the contributions of the previous part, data augmentation and synthetic data are key components of the works in this part. Starting with data augmentation in Self-Supervised Learning (SSL) for computer vision models, the study presented in \Cref{chap:on_the_importance_ssl} provides empirical evidence that data augmentation plays an underestimated role in impacting model performance. No significant improvements were attainable when optimizing training hyperparameters, but significant improvements were achieved when optimizing data augmentation hyperparameters. This finding further motivated the development of the novel data augmentation strategy of Hard View Pretraining (HVP) presented in \Cref{chap:hvp}. HVP leverages meta-learning to select challenging views based on the model’s current performance by extending the random view generation common in SSL. Despite its simplicity, pretraining with HVP results in improved hyperparameter robustness and downstream task performance across architectures, datasets, and SSL methods. HVP also achieves a new state-of-the-art result on the ViT-B/16 transformer architecture and demonstrates the potential of meta-learned data augmentation strategies. Building on this foundation, the work in \Cref{chap:learningses} uses meta-learning to optimize not just data augmentation but the data generation process itself. This work demonstrates the strength of adopting meta-learning for optimizing the effective use of data: the meta-learned synthetic proxies exhibit valuable properties, such as facilitating agent training with significantly fewer training steps and greater robustness to agent hyperparameter variations. However, meta-learning synthetic environments is computationally expensive, even in the relatively controlled domain of RL. To address this, the work presented in \Cref{chap:oswm} investigates reducing the costs by replacing meta-learning with randomly sampled synthetic data. The approach samples the data from network-based prior distributions that mimic the dynamics of multiple RL environments, allowing the learned proxy to support agent training across diverse target environments. By achieving a proxy environment that generalizes to multiple targets, we further reduce computational demands and take a step toward foundational world models trained entirely on synthetic data.

\chapter{Takeaways and Outlook}\label{chap:future}
This dissertation has explored how meta-learning and synthetic data can advance automated pretraining and finetuning. The works throughout this dissertation encompass a broad range of tasks, including image classification and large language model finetuning, pretraining with self-supervised learning (SSL), and world model learning for reinforcement learning (RL). Together, the contributions in this dissertation illustrate that meta-learning can inform strategies for selecting pretrained models and tuning their hyperparameters, automated data augmentation, and synthetic data generation. By integrating insights from diverse domains, this dissertation presents novel approaches that go a step towards more automated, adaptive, and generalizable machine learning workflows.\\

\subsection*{Meta-Learning for Automated Model Selection and Finetuning}
With ZAP, we took initial steps toward adapting the principles of Combined Algorithm Selection and Hyperparameter Optimization (CASH) to modern deep learning pipelines. Crucially, ZAP demonstrated that when combined with meta-features, suitable augmentations, and a large-scale meta-dataset, such AutoML methods can be applied effectively in the deep learning domain to automate finetuning. ZAP introduced a zero-shot approach, where no exploratory evaluations on the target dataset were required for the DL pipeline selection. ZAP’s dataset level augmentation strategy revealed how expanding and diversifying a meta-dataset allows the selection process to accommodate test-time variation and handle missing values. We believe this robustness likely arises from the model’s capacity to capture correlations across pipelines in a geometric space, generalizing beyond observed conditions. Building on ZAP, Quick-Tune proposed a few-shot setting and leveraged partial learning curves to meta-learn probabilistic surrogate models. This integration allowed Quick-Tune to iteratively select pipelines through Bayesian optimization under practical user time and resource constraints.\\

Together, ZAP and Quick-Tune have established a foundation for automated finetuning in the deep learning era. We believe these works established a baseline by showing that tailored model selection and hyperparameter tuning can outperform finetuning large, general-purpose billion-parameter models like DINOv2, especially when the target tasks differ substantially from the models' original pretraining domains. A pivotal design decision was the integration of finetuning and data augmentation strategies in Quick-Tune's search space, aligning with ongoing developments in machine learning practices. The ability of methods like ZAP and Quick-Tune to explore vast search spaces and their compatibility with newly emerging methods like parameter-efficient finetuning (PEFT) \citep{lialin-arxiv23a, xu-arxiv23a} such as LoRA \citep{lora} gives them a competitive edge. Notably, ZAP and Quick-Tune illustrate how CASH can be adopted for a quickly growing space of LLMs. For example, they allow generalizing to LLMs' post-pretraining phase that encompasses a multitude of adaptations such as finetuning, PEFT, adapters, prefix tuning, and chain-of-thought \citep{wei2022chain} design. As PEFT and other technologies become more widespread, automated methods such as ZAP and Quick-Tune may become increasingly relevant as the importance of systematically exploring configuration spaces grows. These methods likewise introduce a potentially large number of additional hyperparameters that require systematic exploration. For instance, to just name a few: low-rank dimension, adapter size, chain-of-thought styles, or number of LLM agents in multi-agent settings. Lastly, as large-scale model training requires large-scale data pipelines, more and more ``support models" are being developed, such as for sample deduplication detection, RLHF rewards, high-quality prompts and responses selection, or even for curriculum learning \citep{bengio2009curriculum, llama-3-3-modelcard} (e.g., when and how to mix real with synthetic data). By unifying domain knowledge with large-scale hyperparameter search and model selection, automated methods like ZAP or Quick-Tune can play a pivotal role in future LLM research and development.\\

However, not all design decisions during the development of ZAP and Quick-Tune were fruitful. For instance, attempts to improve performance by integrating learned meta-features \citep{achille-iccv19a} in both ZAP and Quick-Tune have so far underperformed simpler, static meta-features, which remain insufficiently understood and beg future investigation. Furthermore, while we have shown the applicability of Quick-Tune to the language processing domain for finetuning LLMs, we found that empirical search performance is better when no iterative refinement of the surrogate models is executed. We hypothesize this is due to overfitting to a meta-dataset that is too small, hinting at the disadvantage of few-shot and Bayesian Optimization in contrast to zero-shot approaches. However, more experiments with larger meta-datasets are required to validate this hypothesis and to understand this phenomenon better. Apart from that, a range of open questions and refinements still exists. For instance, meta-hyperparameters of our approaches, such as the computational budget allocated for evaluating candidate configurations, can significantly influence performance and achievable speed-ups over black-box methods. How to encode hyperparameters and architectures remains an underexplored factor, as a default practice is to rely on categorical embeddings \citep{feurer-nips15a} mostly. Future work could also extend these approaches to the domain of large \emph{multimodal} models, and leveraging multi-GPU and distributed computation setups could further broaden their impact and reveal how they respond to automated optimization. Moreover, recent work \citep{liu-iclr24a} proposed using LLMs as surrogate models. This poses an intriguing alternative for the performance and cost prediction of DL pipelines that can extend to multi-objective optimization. Building on this, one could also use LLMs to directly output the kernel function in Bayesian Optimization for pipeline cost and performance prediction. Addressing these points will help move from promising baselines and initial successes toward more mature and widely deployable automated finetuning solutions in deep learning.\\

\subsection*{Meta-Learning Data Augmentation and Synthetic Data for Enhanced Learning}
Beyond addressing CASH in the deep learning context, this dissertation underscores the importance of viewing data augmentation and synthetic data as central, meta-learned elements rather than peripheral components of learning pipelines. Our studies illustrate empirically that carefully leveraging data augmentation and synthetic data can yield significant performance improvements in supervised, self-supervised, and reinforcement learning. For instance, in self-supervised learning, we investigated the role of data augmentation. We showed that even simple, learning-free automated data augmentation techniques like Hard View Pretraining (HVP) proved sufficient to challenge existing discriminative pretraining pipelines and to improve robustness and downstream performance. This empirically shows that advancements in automated data augmentation do not need to rely on parametric or complex architectural changes but can be achieved by effectively leveraging state-dependent augmentation strategies. However, we also noticed that some early design decisions were less successful. For example, training a network to generate augmented views resulted in unstable training, as the network generated either very hard or static views, resulting in repeated model collapse. This shows how challenging it can be to design augmentation strategies that adapt to the learning state without destabilizing a given training process. Going forward, future work could use the HVP objective to learn the parameters of parameterized distributions of data augmentation operations instead of learning the view generation itself. This may not only make the task more computationally affordable than learning entire view generators as is common in the literature but also allow for an interpretation of the required augmentations at any given learning state. Moreover, the HVP paradigm can be applied wherever random augmentations sampling is used, not just in pretraining but also during finetuning. Similarly, future work could also extend selecting hard augmentations to multimodal model training such as CLIP \citep{radford-icml21a}. Lastly, HVP's principles could also be adapted for LLM training. In the supervised finetuning phase of LLMs, for example, one could automatically select high-loss samples from each training batch where the model's predictions diverge most from ground-truth labels.\\

With Synthetic Environments (SEs), we turned our attention to the possibility of meta-learning synthetic data itself. SEs allowed agent training without continuous reliance on real-world interactions while drastically reducing the environment interactions needed and increasing agent hyperparameter robustness. Due to their efficiency, agent hyperparameter robustness, and agent algorithm agnosticism, such environment proxies could aid automated reinforcement learning (AutoRL) \citep{parkerholder-jair22a} by enabling the pretraining of general agents or facilitating efficient development cycles for new RL algorithms. Developing analytical tools based on SEs can also facilitate inverse reinforcement learning, providing deeper insights into environment dynamics and task structures. These tools aid human understanding and analysis of complex RL tasks from optimization and parametric perspectives.

The SE framework could also be applicable in the LLM context: current LLM reward models (RMs) often optimize proxy objectives (e.g. human preferences or outcome and rule-based rewards \citep{uesato2022solving}, i.e. they optimize ``what responses are preferred by humans?" instead of ``how to solve tasks?") that lead to LLMs exploiting shortcuts (e.g. poor readability and language mixing) or complex multi-stage training that requires post-rejection sampling to generate synthetic datasets with strong samples for finetuning \citep{guo2025deepseek}. This is because coming up with RMs that facilitate open-ended problem-solving is very challenging. Learned environments and RMs may present an interesting step towards aligning better with open-ended problem-solving capabilities that is worth exploring. The SE framework may offer a pathway to address this by meta-learning task-aligned reward functions through bi-level optimization. For instance, inspired by SEs' ability to meta-learn synthetic environment dynamics and rewards, one could replace the traditional RL agent in the SE framework with an LLM agent in a language task environment (e.g., constrained code generation or math-solving) and meta-learn RMs that elicit task solving behaviors. Moreover, RNs' potential-based reward shaping could ground these rewards in verifiable outcomes (e.g., code executability or answer consistency) and avoid reliance on brittle human judgments or mitigate exploiting reward shortcuts.\\


In the context of world models, the large-scale world model Genie 2 \citep{parker-holder2024genie2} was represented recently which is a first step toward foundational world models that may enable generalist capabilities for open-ended agent learning \citep{wang2019paired}. As architectures and data scale, it may become relevant to incorporate insights from approaches like SEs that exploit algorithmic properties such as agent agnosticism, hyperparameter robustness, and training efficiency. For instance, agents could be trained using Genie 2's pixel space, while SEs operate in the latent embedding space to model the underlying dynamics and actions. The SEs could be meta-learned by adjusting their dynamics predictions based on the agent’s performance on a target task. Combining such latent-space modeling with Genie 2 may provide a pathway to integrating meta-learning approaches like SEs with AutoRL and foundation world models.

Motivated by world models that model multiple tasks or target environments, we introduced the One-Shot World Model (OSWM). OSWM leverages randomly sampled data from mixed prior distributions and in-context learning to pretrain a world model that approximates the dynamics of multiple simple control environments. A key insight is that the complexity and diversity of the prior distribution correlate positively with the performance of the learned agent policy, emphasizing the importance of designing sophisticated priors. While still very limited, OSWM demonstrates what may be possible by advancing our understanding of how prior distributions should be modeled. With well-designed priors tailored to target tasks, agents and robots could be pretrained in simulation before deployment, potentially reducing risks such as failure modes that damage robots and reducing costs by requiring less real-world data. This approach offers a path toward safer and more efficient pretraining strategies for real-world applications.\\

All in all, this dissertation has contributed to the development of individual components and solutions that advance the vision of fully automated, meta-learned machine learning systems for pretraining and finetuning. By combining the automated model selection and hyperparameter optimization capabilities of ZAP, Quick-Tune, and Quick-Tune-Tool jointly with the data augmentation and synthetic data strategies provided by HVP, GroupAugment, SEs, and OSWM, future research can develop unified systems that integrate components like these and further automate key elements of the machine learning pipelines. As computational resources grow, these components could be meta-learned jointly to unlock synergies beyond isolated automation. Such systems would tailor solutions to specific tasks and domains by dynamically selecting high-performing configurations, adjusting data strategies, and integrating synthetic environments. The modularity of this dissertation's contributions allows for refining components individually or combining them to enhance their performance and address the evolving complexities of modern AI applications. Lastly, the contributions in this dissertation interface with data in a manner that imposes no strong assumptions about its representation, ensuring that these methods can seamlessly extend to new model architectures, including LLMs and future model architectures. This broad applicability positions our meta-learned solutions as versatile candidates for future machine learning deployment and research.

\part*{Appendices}
\addcontentsline{toc}{part}{Appendices}
\appendix

\chapter{Appendices for Zero-Shot AutoML with Pretrained Models}\label{appendix:sup_zap}
\section{Paper Appendix}
\includepdf[pages=14-17]{chapters/papers/2022_zap_icml.pdf}
\section{Statement of Contributions}
\thispagestyle{empty}
\includepdf[pages=1-5]{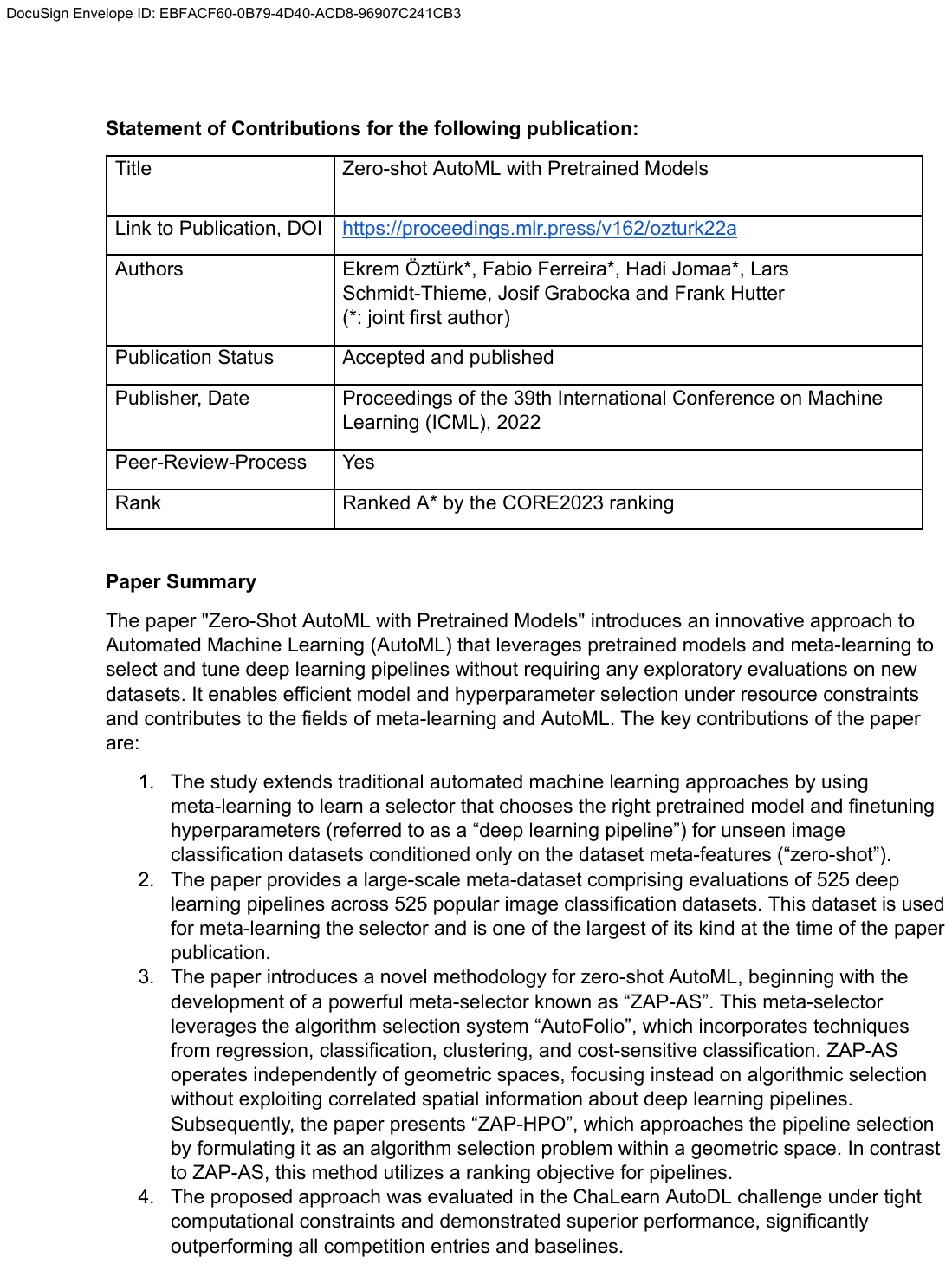}

\chapter{Appendices for Quick-Tune: Quickly Learning Which Pretrained Model to Finetune and How}\label{appendix:sup_qt}
\section{Paper Appendix}
\includepdf[pages=15-]    {chapters/papers/2024_quicktune_iclr.pdf}
\section{Statement of Contributions}
\thispagestyle{empty} 
\includepdf[pages=1-4,scale=1.0,offset=0 -1.25cm]{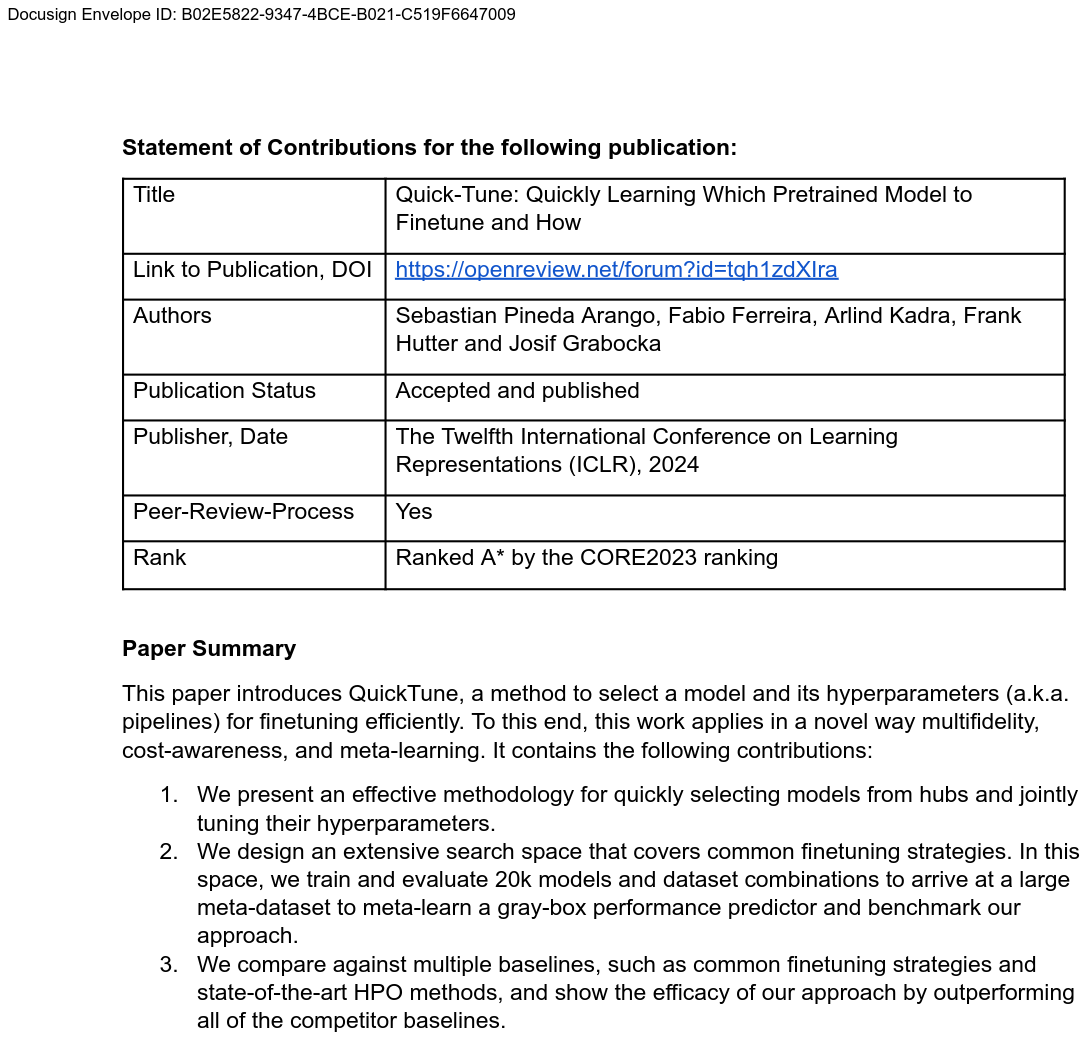}

\chapter{Appendices for Quick-Tune-Tool: A Practical Tool and its User Guide for Automatically Finetuning Pretrained Models}\label{appendix:sup_qtt}
\section{Paper Appendix}
\includepdf[pages=11-14]{chapters/papers/2024_qtt_automl.pdf}
\section{Statement of Contributions}
\thispagestyle{empty}
\includepdf[pages=1-4,scale=1.0,offset=0 -1.25cm]{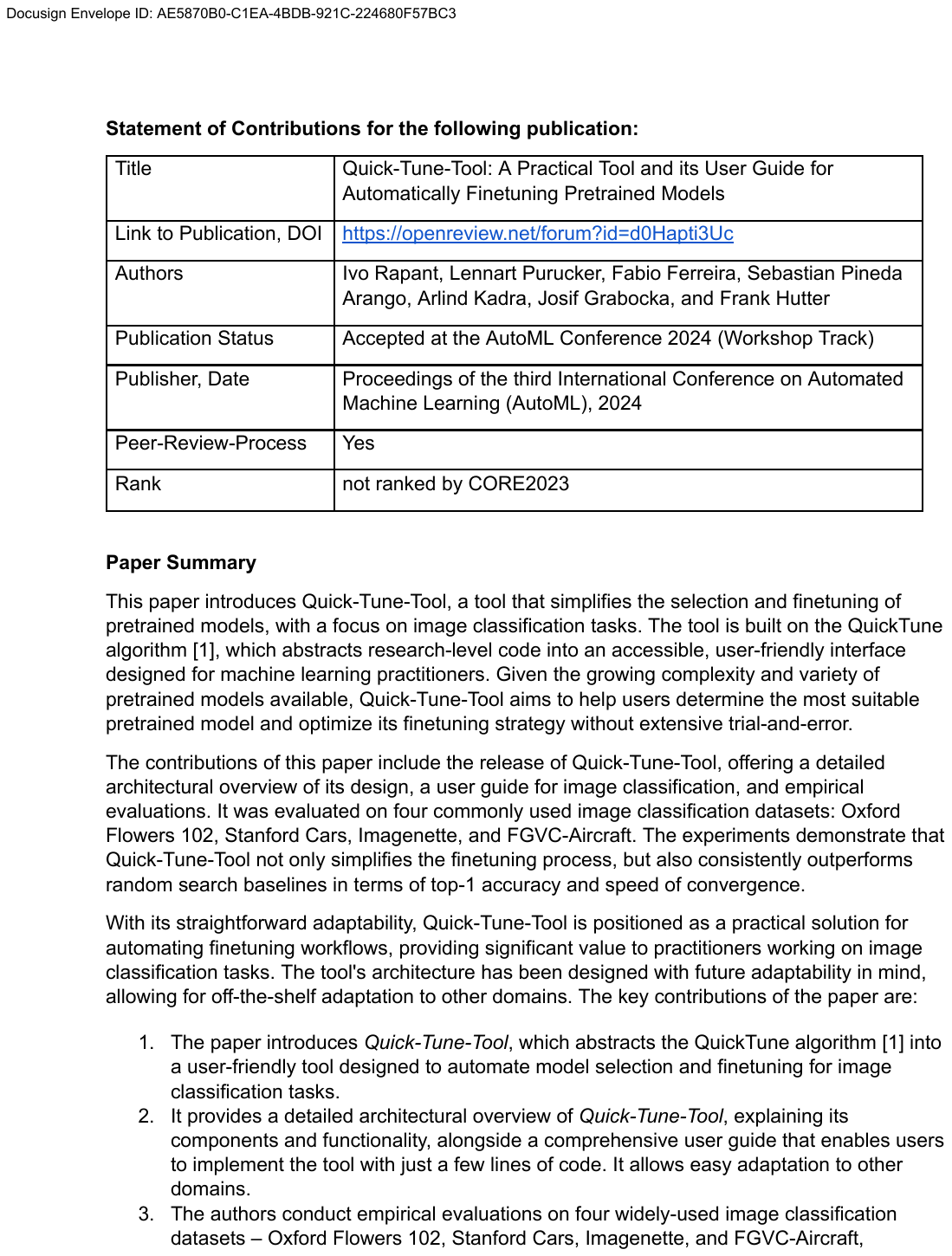}

\chapter{Appendices for Transfer Learning for Finetuning Large Language Models}\label{appendix:sup_qt_llms}
\section{Paper Appendix}
\includepdf[pages=7-12]{chapters/papers/2024_quicktune_llms_neurips.pdf}
\section{Statement of Contributions}
\thispagestyle{empty}
\includepdf[pages=1-3,scale=1.0,offset=0 -1.25cm]{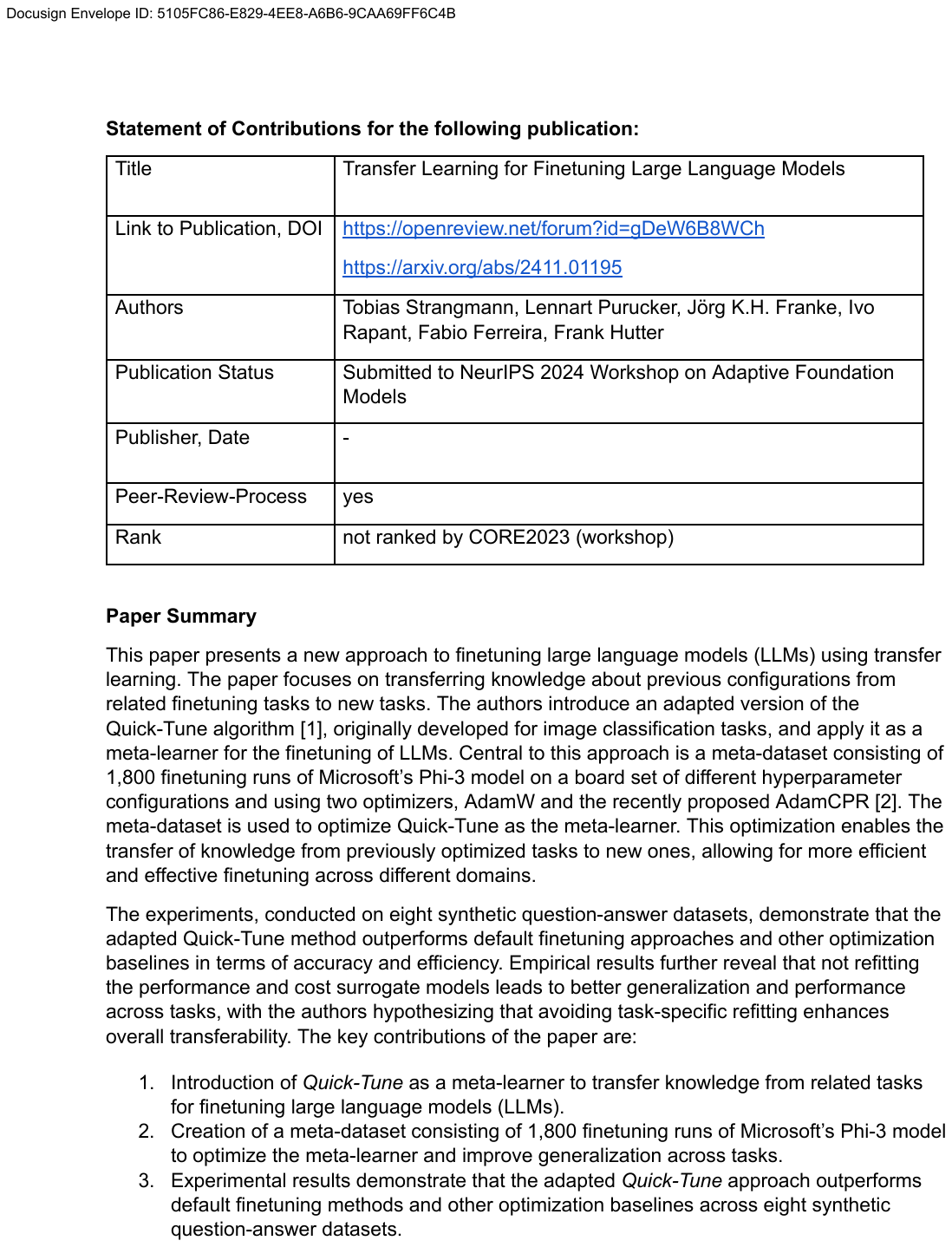}

\chapter{Appendices for Beyond Random Augmentations: Pretraining with Hard Views}\label{appendix:sup_hvp}
\section{Paper Appendix}
\includepdf[pages=13-]   {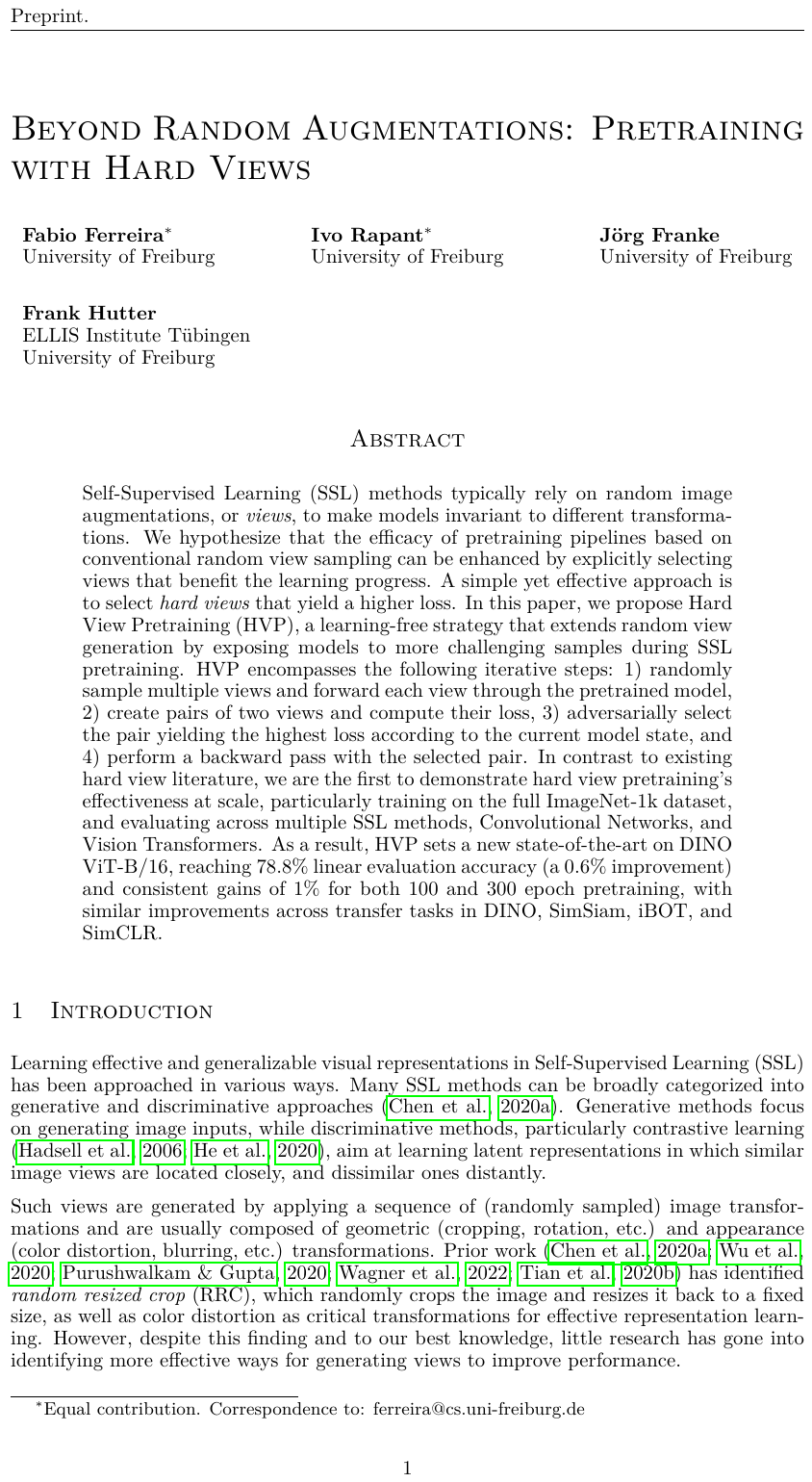}

\section{Statement of Contributions}
\thispagestyle{empty}
\includepdf[pages=1-4,scale=1.0,offset=0 -1.25cm]{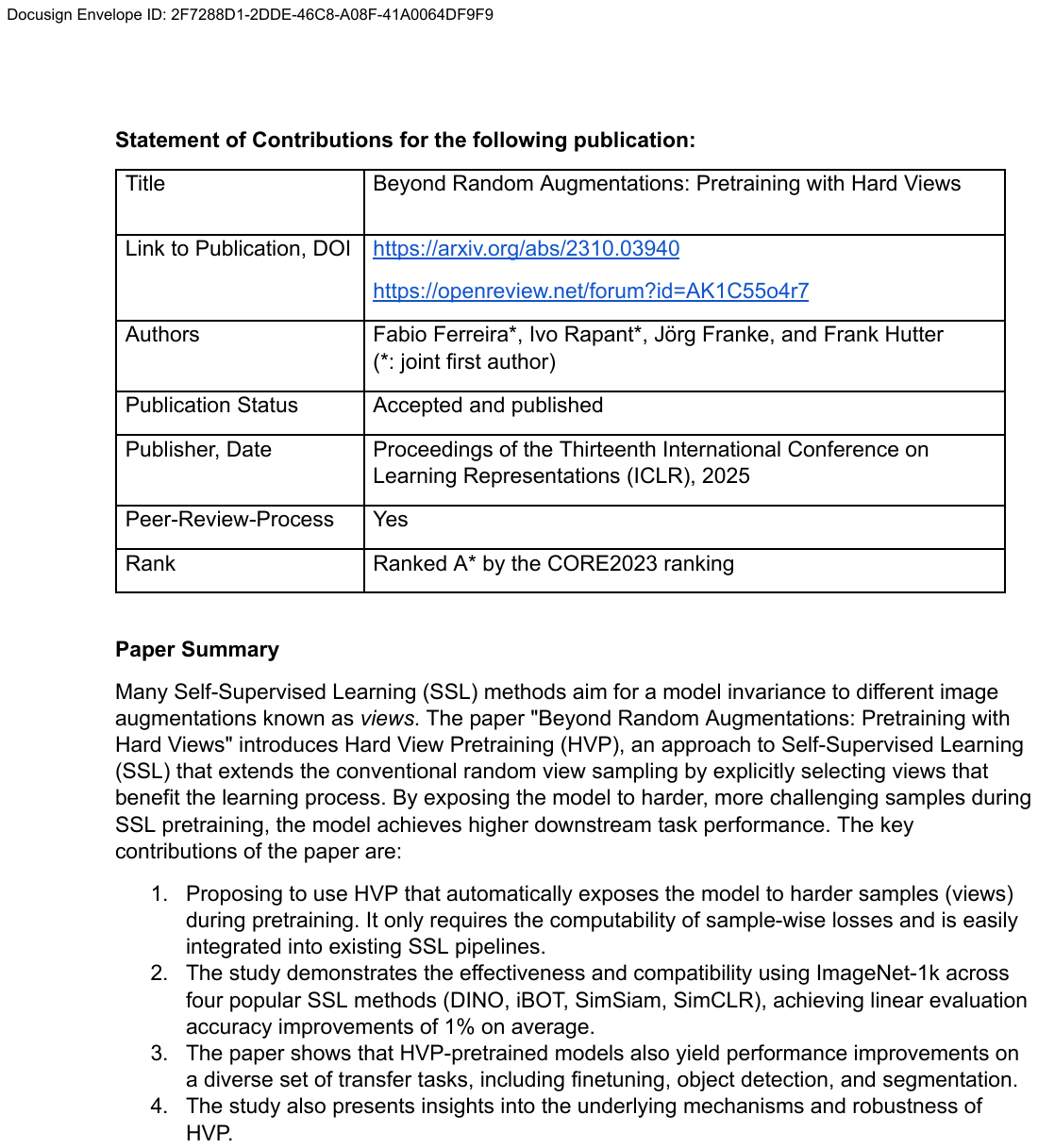}

\chapter{Appendices for On the Importance of Hyperparameters and Data Augmentation for Self-Supervised Learning}\label{appendix:sup_on_importance_ssl}
\section{Paper Appendix}
\includepdf[pages=7-]{chapters/papers/2022_onimportance_neurips.pdf}
\section{Statement of Contributions}
\thispagestyle{empty}
\includepdf[pages=1-4,scale=1.0,offset=0 -1.25cm]{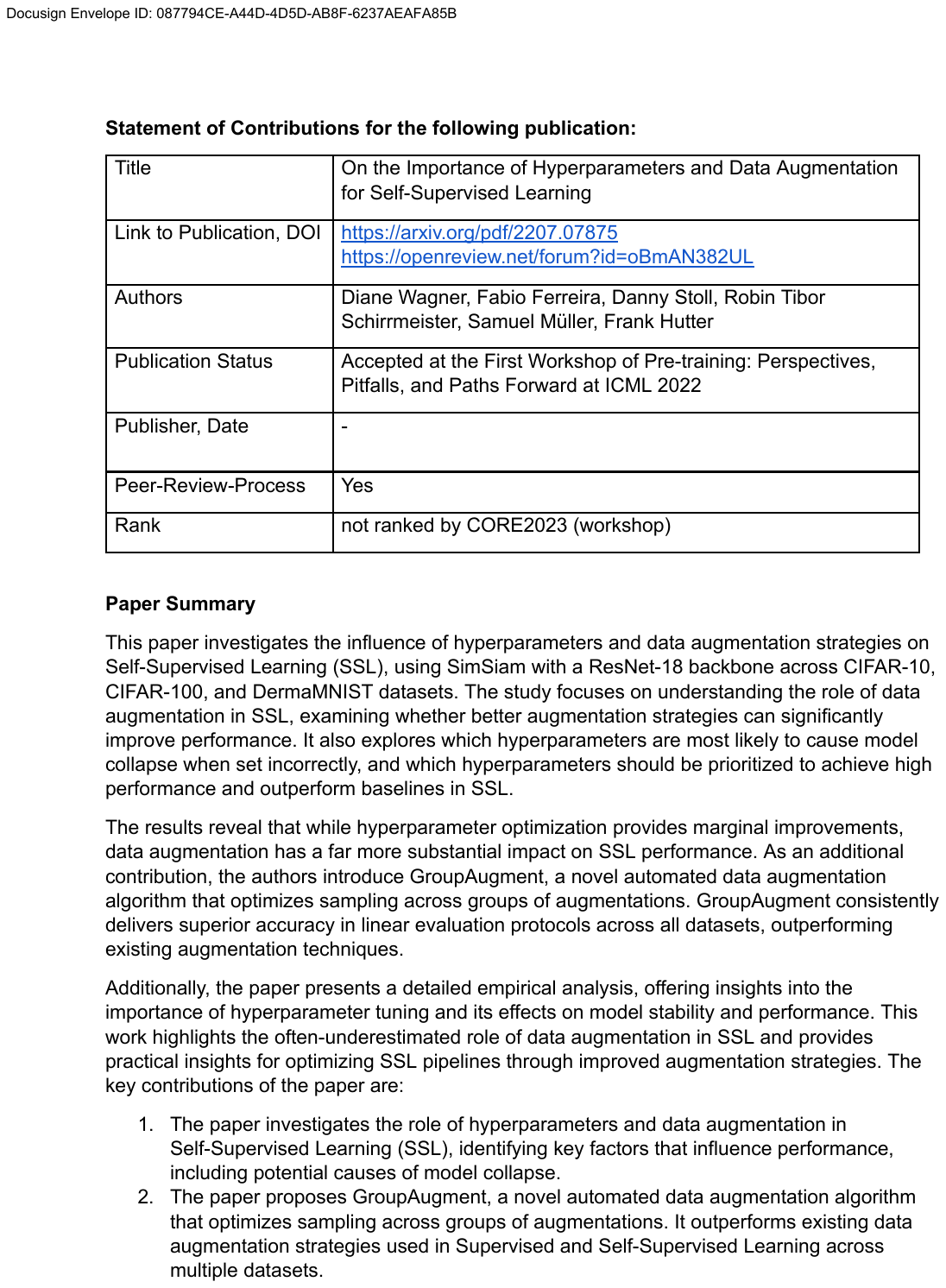}

\chapter{Appendices for Learning Synthetic Environments and Reward Networks for Reinforcement Learning}\label{appendix:sup_se}
\section{Paper Appendix}
\includepdf[pages=14-]{chapters/papers/2022_learningses_iclr.pdf}
\section{Statement of Contributions}
\thispagestyle{empty}
\includepdf[pages=-,scale=1.0,offset=0 -1.25cm]{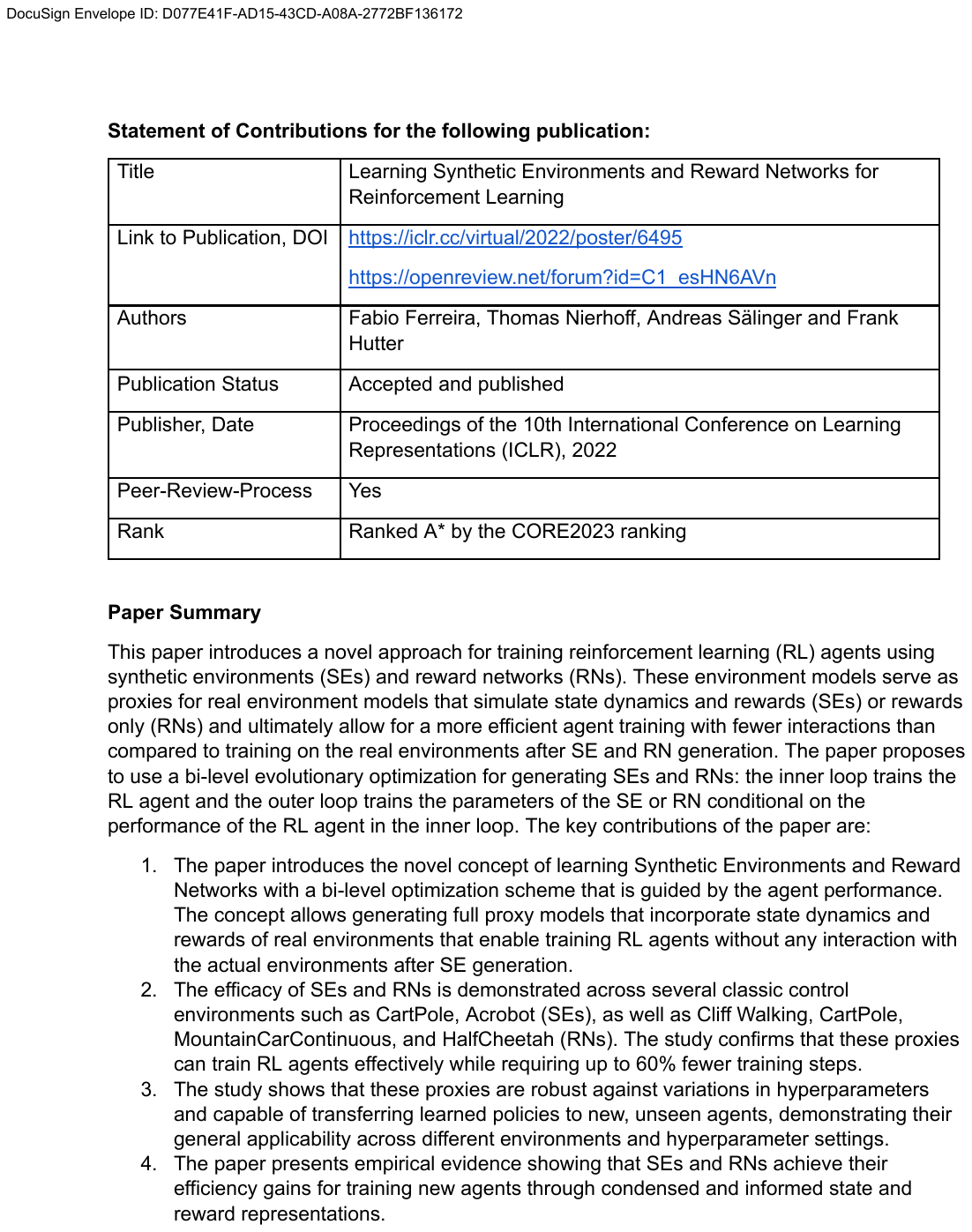}

\chapter{Appendices for One-shot World Models Using a Transformer Trained on a Synthetic Prior}\label{appendix:sup_oswm}
\section{Paper Appendix}
\includepdf[pages=12-]{chapters/papers/2024_oswm_neurips}
\section{Statement of Contributions}
\thispagestyle{empty}
\includepdf[pages=-,scale=1.0,offset=0 -1.25cm]{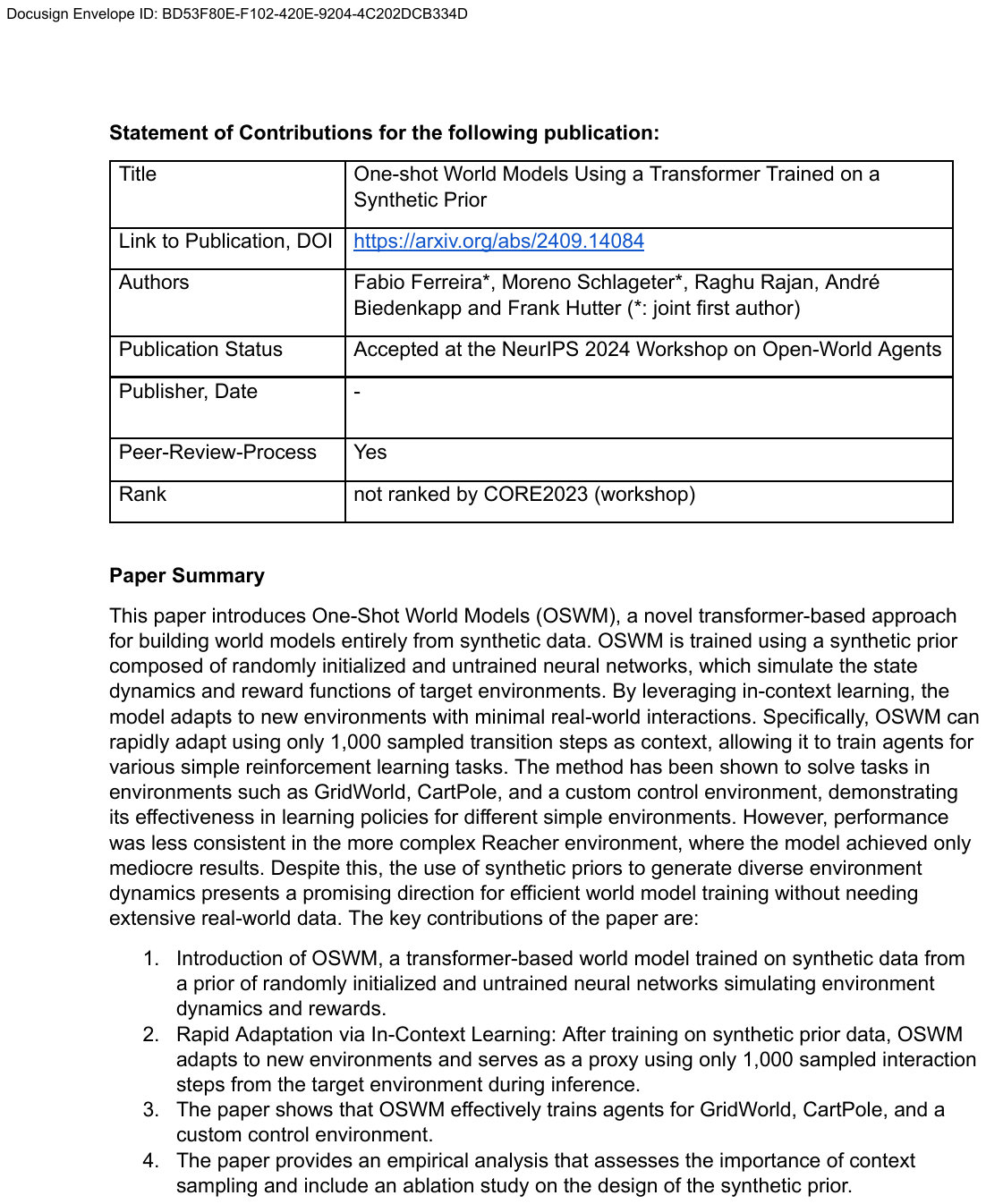}
\backmatter

\part*{Bibliography}
\addcontentsline{toc}{part}{Bibliography}
\markboth{Bibliography}{}
\setlength\bibitemsep{1.5\itemsep}
\printbibliography[heading=none,notcategory=exclude]

%

\end{document}